%% file: main.tex
\documentclass[preprint,authoryear,nonatbib]{elsarticle}

\usepackage{amssymb}
\usepackage{amsmath}
\usepackage{enumitem}
\usepackage[american]{babel}
\usepackage{csquotes}
\usepackage[
  backend=biber,
  style=apa,
  natbib=true,
  doi=true,
  url=true
]{biblatex}
\usepackage{hyperref}
\hypersetup{hidelinks}

\usepackage{xcolor} 

\makeatletter
\def\ps@pprintTitle{%
  \let\@oddhead\@empty
  \let\@evenhead\@empty
  \def\@oddfoot{\hfil\thepage\hfil}%
  \let\@evenfoot\@oddfoot
}
\makeatother

\begin{document}

\begin{frontmatter}



\title{Historical Anchors for Scientific Forecasting on Human-Aware Knowledge Graphs}

\author[1]{Siu Tin Wu}
\ead{stwu@lanl.gov}
\author[2]{Ryan C. Barron}
\ead{barron@lanl.gov}
\author[2]{Phoebe Upton Sanderbeck}
\ead{phoebeu@lanl.gov}
\author[3]{Kim {\O.} Rasmussen}
\ead{kor@lanl.gov}
\author[4]{George Tompkins}
\ead{tompkins@lanl.gov}
\author[2]{Maksim E. Eren\corref{cor1}}
\ead{maksim@lanl.gov}

\cortext[cor1]{Corresponding author}

\affiliation[1]{organization={Advanced Research in Cyber System, Los Alamos National Laboratory},
            city={Los Alamos},
            state={New Mexico},
            country={USA}}

\affiliation[2]{organization={Computational Intelligence \& Modeling, Los Alamos National Laboratory},
            city={Los Alamos},
            state={New Mexico},
            country={USA}}
            
\affiliation[3]{organization={Fluid Dynamics and Solid Mechanics, Los Alamos National Laboratory},
            city={Los Alamos},
            state={New Mexico},
            country={USA}}

\affiliation[4]{organization={Intelligence \& Systems Analysis, Los Alamos National Laboratory},
            city={Los Alamos},
            state={New Mexico},
            country={USA}}

\begin{abstract}
\input{sections/abstract}
\end{abstract}



\begin{keyword}
Artificial Intelligence \sep Science Forecasting \sep Knowledge Graphs \sep Graph Embeddings \sep Random Walks



\end{keyword}

\end{frontmatter}



\section{Introduction}
\input{sections/intro}

\begin{figure*}[!t]
    \centering
    \includegraphics[width=\textwidth,keepaspectratio,trim=0in 0in 0in 1.8in,clip]{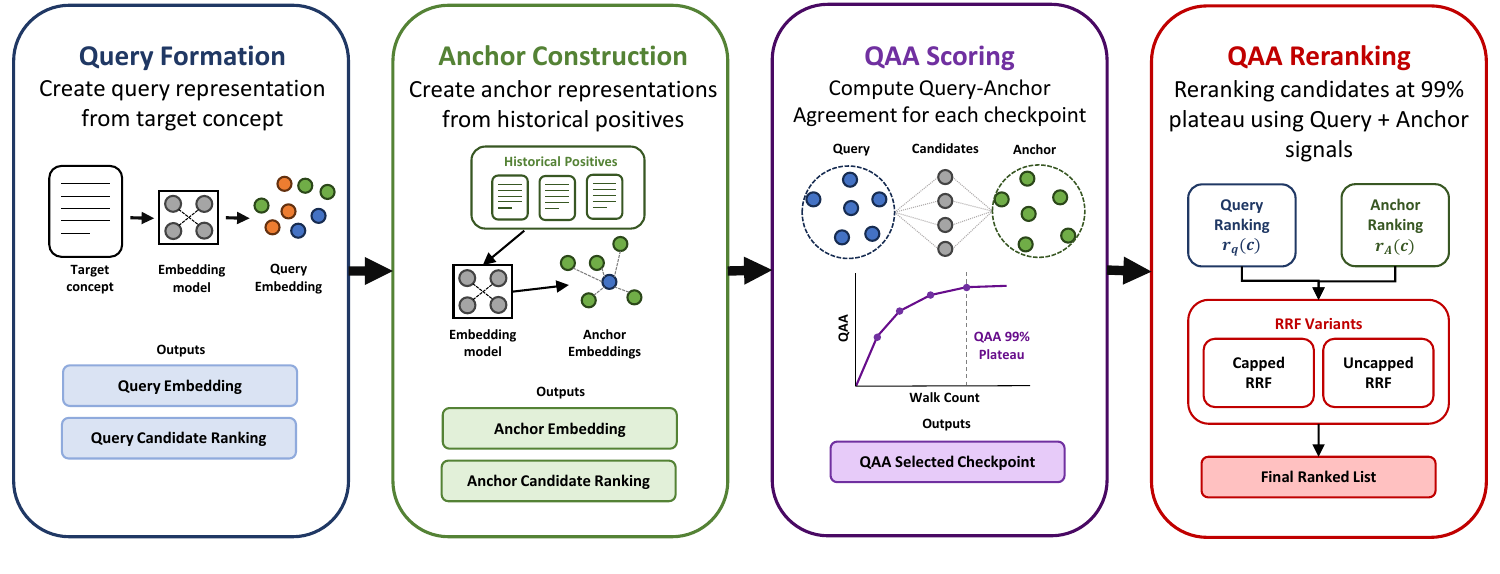}
    \caption{Overview of the Query--Anchor Agreement (QAA) workflow. \textbf{Query Formation:} the target concept is used as a query and an embedding model is trained using DeepWalk. Candidates are ranked by input--output cosine similarity to the query in the learned embedding. \textbf{Anchor Construction:} entities that represent known positives (materials known to be thermoelectric or drugs associated with the \emph{coronavirus} property) are collected as anchors. Each candidate is scored by its average input--input similarity to its $k_A=10$ nearest anchors, giving us our anchor-based ranking. \textbf{QAA Scoring:} the query-based and anchor-based rankings are compared by their top-window overlap, averaged over various window sizes, to produce a single agreement score (QAA). We smooth the QAA score over the walks and select the earliest saturation point, where QAA is 99\% of its highest value for our QAA selected checkpoint. \textbf{QAA Reranking:} within the selected checkpoint, the query and anchor rankings are fused by reciprocal rank fusion---either capped (anchors only support query-favored candidates) or uncapped (anchors given full influence)---to refine the final candidate ordering.}
    \label{fig:qaa_workflow}
\end{figure*}

\section{Related Work}
\input{sections/related_work}

\section{Background}
\label{sec:background}
\input{sections/background}

\section{Method: Query--Anchor Agreement}
\label{sec:qaa}
\input{sections/methods}

\section{Experimental Setup}
\label{sec:experimental_setup}

\input{sections/experimental_setup}

\section{Results}
\input{sections/results}

\section{Discussion and Future Work}
\input{sections/discussions}

\section{Conclusion}
\input{sections/conclusion}

\section*{CRediT authorship contribution statement}
\input{sections/credit}

\section*{Declaration of competing interest}
\input{sections/competing_interests}

\section*{Data availability}
\input{sections/data_availability}

\section*{Acknowledgment}
\input{sections/acknowledgment}

\section*{Declaration of generative AI and AI-assisted technologies\\in the manuscript preparation process}
\input{sections/ai_declaration}



\printbibliography[title={References}]

\end{document}

%% file: sections/abstract.tex
Predicting which scientific connections will form next can point researchers toward the most promising leads. One successful approach learns representations by sampling paths through publication networks, represented by a hypergraph, that connect scientific entities, concepts, and researchers, approximating how knowledge and expertise move through the literature. Sampled paths are then used to train an embedding model to forecast scientific connections such as a property of a material. Performance, however, depends strongly on how many paths are sampled: too little sampling yields incomplete representations, while further sampling can eventually weaken prediction quality. The best sampling budget is therefore often identifiable only after future discoveries are known. We introduce \textbf{Query--Anchor Agreement} (QAA), a diagnostic requiring no future labels or separately trained model. QAA can be used to select a sampling budget, and refine or re-rank the resulting forecast to improve forecasting scientific connections. QAA compares the candidates a model predicts for a target with those most similar to the target's known historical examples, and identifies when these two views converge. Across thermoelectric-material and COVID-19 drug-repurposing benchmarks, repeated random-seed runs, and three sampling policies, QAA-selected budgets fall on average 1.86 percentage points below hindsight-optimal selection for broad candidate recovery. Further, promoting candidates supported by both views, under a capped fusion that bounds the anchor contribution, improves broad recovery across all six benchmark-policy settings, which define the types of network walks used for sampling.

%% file: sections/intro.tex
Forecasting which scientific connections will form next can help direct scarce scientific effort toward promising materials, drugs, and hypotheses. This possibility has been demonstrated in several settings: unsupervised embeddings trained on materials-science abstracts can recover latent structure--property relationships and recommend materials years before their discovery~\citep{tshitoyan2019unsupervised}, while \citet{sourati2023accelerating} show incorporating researcher and author information further improves forecasting of publication-derived network models. These approaches rest on a common idea. The scientific literature can be represented as a network in which publications connect their authors, the materials or drugs they study, and the concepts they mention. 

In Sourati and Evans' implementation, their framework samples paths, or random walks, through this network, approximating how knowledge and expertise move across the literature~\citep{sourati2023accelerating}. Then, an embedding model learns from the sampled network paths, placing entities that appear in similar contexts close together. For a target concept, such as thermoelectricity or coronavirus, candidates such as materials for the former or drugs for the latter are ranked by their proximity to the target based on the distance in the embedding space, and highly ranked candidates are treated as plausible future connections. Two choices shape the resulting forecast: the walk policy, which determines the kinds of paths through the network that the model favors, and the sampling budget, which determines how many paths it observes. Some policies preserve author-mediated routes between fields, whereas others remain closer to entity-to-entity connections (Section~\ref{sec:background}).

Both the walk-policy and number of walks for sampling choices affect forecast quality, but this work focuses on selecting a sampling budget within a chosen walk policy. Too little sampling may provide an incomplete view of the target-relevant network (under-fitting), yet additional sampling does not reliably improve the forecast: rankings can improve, plateau, or decline as more walks are added (over-fitting). The best budget is therefore typically identifiable only after future discoveries are known, when the model's performance can be tested~\citep{antoniak2018evaluating,wendlandt2018factors,krenn2023forecasting}. Such retrospective selection is unavailable when a forecast must be made prospectively. Recovering previously known examples is an inadequate replacement because it rewards reproduction of the historical target neighborhood, which may be dense and well studied, rather than the identification of connections that remain undiscovered. What is missing is a forecast-time signal for determining whether the representation has reached a stable, target-relevant sampling regime.

In this paper, we expand upon Sourati and Evans' implementation~\citep{sourati2023accelerating}, and introduce \textbf{Query--Anchor Agreement} (QAA), a forecast-time diagnostic that requires no future labels or separately trained model. For each sampling budget, QAA compares two rankings of the same candidate pool: one produced by the target concept and another produced by entities associated with the target, or a closely related concept, before the forecast date. We call these historical examples \emph{anchors}; in thermoelectric forecasting, for example, the anchors are materials already known to be thermoelectric. Rather than treating anchors as surrogate labels for forecast accuracy, QAA uses them as an independent structural reference. When the target and its anchors increasingly point to the same candidates and their agreement saturates, QAA identifies a target-relevant sampling regime within the chosen walk policy. The same anchor signal can then refine the forecast by promoting candidates supported by both the target and its historical examples, providing a reranked final list.

We evaluate QAA on the released thermoelectric-material and COVID-19 drug-repurposing benchmarks of \citet{sourati2023accelerating}, preserving their prediction tasks, candidate pools, prediction dates, and walk policies while reserving future discovery labels for evaluation only. Dense sweeps over walk budgets, policies, and random seeds reveal substantial configuration sensitivity and confirm that additional sampling is not monotonically beneficial, so a fixed walk budget is not a one-size-fits-all choice: the useful sampling regime shifts with the benchmark, the walk policy, and the structure of the underlying graph. Across these sweeps, QAA selects budgets that come close to the best achievable in hindsight for broader candidate recovery, without consulting any future labels. Reusing the same anchor signal to refine the ranking of the predictions further improves broad recovery across benchmarks and policies, in several cases surpassing even the retrospectively best forecast. Together, these results provide a practical path toward assessing and refining scientific discovery forecasts using only information available when the forecast is made.

%% file: sections/related_work.tex
Forecasting scientific discovery from the research literature has been demonstrated using content-only embeddings. Unsupervised models trained on materials-science text can recover latent structure--property relationships and recommend materials before their later discovery~\citep{tshitoyan2019unsupervised}. \citet{sourati2023accelerating} extend this approach by incorporating researcher activity into the forecasting model, demonstrating that human-aware representations can further improve discovery prediction.

A parallel line of work forecasts future science from concept graphs, in which nodes represent scientific concepts and edges indicate that two concepts were studied together in a paper. Early work in this direction predicts future concept connections from semantic networks built over the historical physics literature~\citep{krenn2020trends}. \citet{krenn2023forecasting} introduce the Science4Cast benchmark for predicting future links in an evolving network of AI concepts and report that supervised classifiers using engineered network features---quantities manually derived from the graph rather than learned by an embedding model---outperform end-to-end embedding methods. This pattern persists in subsequent work: forecasts of the citation impact of previously unstudied concept pairs rely entirely on engineered features~\citep{gu2025forecasting}, while DELPHI~\citep{weis2021learning} retains node2vec embeddings only as an auxiliary feature when they improve performance. Together, these results suggest that embedding-based representations cannot be assumed to provide reliable forecasts on their own, motivating methods such as ours that assess their consistency before using them to rank candidate discoveries.

Related approaches forecast other scientific outcomes rather than concrete future links. AUGUR~\citep{salatino2018augur}, for example, predicts the emergence of new research topics from topic co-occurrence networks. This objective also connects to the literature-based discovery tradition, which infers latent relationships from disconnected bodies of work~\citep{swanson1986fishoil,swanson1997complementary}. Later systems operationalize this idea as automatic hypothesis generation over literature-derived networks~\citep{sybrandt2017moliere}, including temporal formulations that predict the future connectivity of concept pairs in evolving publication graphs~\citep{akujuobi2022tpair,zhou2022tde,shariatmadari2025conceptdrift}. Across these methods, the primary output is a predictive score or warning concerning a future event, such as link formation, citation impact, or topic emergence. QAA serves a fundamentally different diagnostic role: rather than estimating the outcome itself, it tests whether the learned representation has stabilized sufficiently to support forecasting, after which our reranking procedure refines the ordering of candidate discoveries.

This ranked-candidate formulation is rooted in the classical link-prediction problem introduced by \citet{libennowell2007link}. Their framework ranks candidate node pairs from a network snapshot using topological proximity measures such as common neighbors, Adamic/Adar, Katz, and rooted PageRank, with performance evaluated using top-$n$ precision. Our setting retains this ranked-candidate structure but replaces fixed proximity heuristics with random-walk embeddings learned from publication-derived hypergraphs~\citep{sourati2023accelerating}, following the broader tradition of random-walk representation learning on heterogeneous networks~\citep{dong2017metapath2vec}. Their top-$n$ evaluation is therefore the classical analogue of our CP@50 and CP@250 metrics, and we treat this work as foundational framing. Comparing and combining such ranked lists is itself well studied, from formal measures of agreement between truncated top-$k$ rankings~\citep{fagin2003topk} to rank-aggregation methods for fusing multiple orderings~\citep{dwork2001rank}; QAA builds on this by treating the agreement between two candidate rankings as a diagnostic in its own right.

Beyond general link prediction, another research lineage shows that discovery signals are encoded in the local geometry of learned scientific representations, reflecting the neighborhood premise underlying content-only materials embeddings~\citep{tshitoyan2019unsupervised}. This paradigm remains active across scientific domains. Candidate materials can be retrieved for thermoelectric discovery by their embedding proximity to a query material~\citep{qu2024leveraging}, and, in a supervised and prospective setting, classifiers trained on literature embeddings can forecast protein-kinase--cancer relevance years in advance~\citep{ravanmehr2021supervised}. QAA adopts the same neighborhood premise but uses it as a forecast-time consistency check. Specifically, it asks whether query-centered and anchor-centered neighborhoods agree within the same representation, rather than treating the embedding as a fixed and inherently reliable input.

Such a diagnostic is important because embedding neighborhoods are known to be unstable. Nearest-neighbor relationships can shift under small changes to the training corpus~\citep{antoniak2018evaluating}, and embeddings remain unstable even for moderately frequent words, making top-$k$ nearest-neighbor overlap a standard stability measure~\citep{wendlandt2018factors,hellrich2016badcompany}. The same concern arises in graph and network embeddings. Node2vec-based predictions show high variance and substantial hyperparameter sensitivity~\citep{krenn2023forecasting,hacker2022node2vec}; stability is often defined through the invariance of nearest neighbors across runs~\citep{wang2022networkinstability}; and randomness can leave aggregate accuracy stable while individual predictions still vary~\citep{schumacher2022stability}. This has motivated criteria that assess representation quality before downstream labels are available, whether by predicting downstream instability from the embedding itself~\citep{leszczynski2020instability} or by scoring representation adequacy in an unsupervised fashion~\citep{garrido2023rankme}. In the same spirit, small changes in random-walk sampling or embedding training can alter the neighborhoods from which scientific candidates are retrieved. QAA addresses this potential vulnerability directly by evaluating neighborhood agreement at forecast time before the representation is used to rank candidate discoveries.

%% file: sections/background.tex
\begin{figure*}[!t]
    \centering
    \includegraphics[width=\textwidth,keepaspectratio,trim=0in .3in 0in 0in,clip]{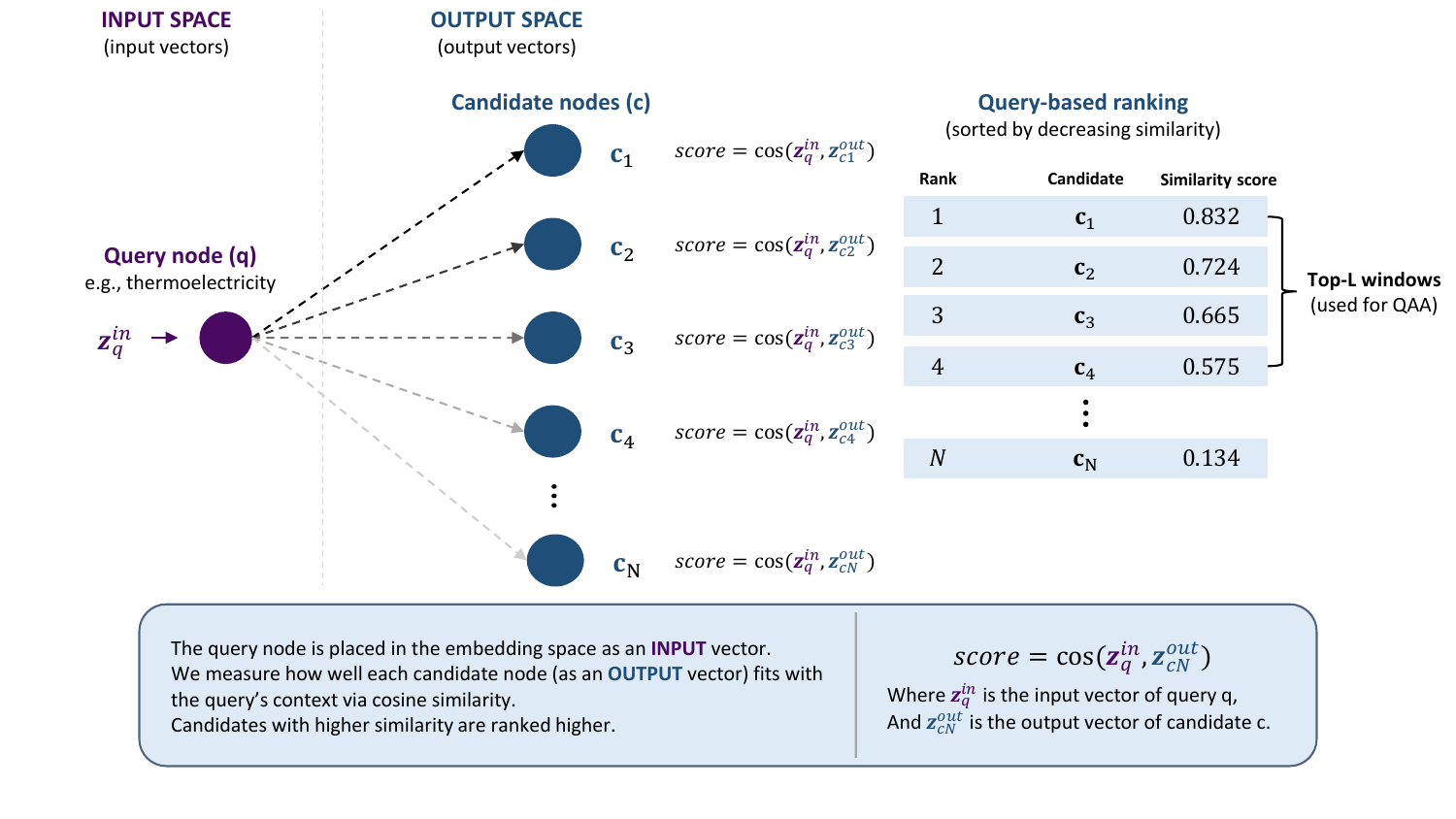}
    \caption{Query-based candidate ranking. The target query is represented by its input vector, each candidate is represented by its output vector, and candidates are ranked by decreasing input-output cosine similarity to the query. The resulting top windows provide the query-centered candidate sets used by QAA.}
    \label{fig:query_ranking}
\end{figure*}

In this work, we expand on the forecasting framework introduced by \citet{sourati2023accelerating}. For completeness, in this section we briefly summarize their approach. Sourati and Evans' framework starts from a scientific corpus and represents each publication as a hyperedge connecting the paper's authors, candidate entities such as materials or drugs, and scientific properties or concepts mentioned in the text. This structure allows discovery forecasts to use more than direct entity--property similarity: candidates can also be connected to a target through shared publications, neighboring concepts, and author-mediated paths through the literature.

Forecasts are generated by sampling random walks over this publication-derived hypergraph, which are then used to learn node embeddings. After training, a target concept is used as a query, and candidate materials or drugs are ranked by embedding proximity to that query. This human-aware framework leverages publication structure and author-mediated pathways as predictive signal, and improves retrospective prediction of future discoveries over content-only baselines.

Their framework also introduces policies governing how walks move through the hypergraph, using \(\alpha\)-modified sampling. Sourati and Evans define three traversal regimes:

\begin{itemize}
    \item \textbf{Uniform} $(\alpha=\mathrm{None})$: samples uniformly from valid non-current nodes in the selected hyperedge.
    \item \textbf{Alpha-1} $(\alpha=1)$: uses balanced group-level sampling over node types, so material/drug nodes and author/property nodes receive comparable group-level probability when both groups are present.
    \item \textbf{Alpha-inf} $(\alpha\to\infty)$: preferentially samples material or drug nodes when possible, suppressing author-mediated traversal and producing a more entity-focused walk policy.
\end{itemize}

Uniform and alpha-1 are human-mediated because walks can pass through author nodes and therefore encode author-mediated exposure, whereas alpha-inf serves as an entity-focused contrast for evaluating the role of human-mediated paths.

These embeddings are learned through a DeepWalk-style procedure~\citep{perozzi2014deepwalk,grover2016node2vec}: Word2Vec skip-gram embeddings~\citep{mikolov2013efficient,mikolov2013distributed} are trained on the generated walk corpora, treating each walk as a sentence-like sequence of graph nodes, so nodes appearing in similar walk contexts receive nearby embeddings. Author tokens are stripped from the walk sequences before training, so authors are never ranked as candidates even though their co-authorship paths still shape the learned neighborhoods.

%% file: sections/methods.tex
Query--Anchor Agreement (QAA) measures whether a trained model is coherent enough to trust at the time a forecast is made, by checking whether the model's top picks for a target agree with the target's known history. Concretely, QAA compares two rankings of the same trained embedding: one ordered by proximity to the target query, and one ordered by proximity to the query's known pre-prediction positives, or \emph{anchors}. The following subsections define the query-based ranking (Section~\ref{sec:query_ranking}), the anchor set and anchor-induced ranking (Sections~\ref{sec:anchor_construction}--\ref{sec:anchor_ranking}), and the multiscale agreement measure and its use as a forecast-time saturation signal (Sections~\ref{sec:multiscale_qaa}--\ref{sec:qaa_saturation}).

QAA builds on the scientific-embedding premise that credible candidates extend a target's known neighborhood~\citep{tshitoyan2019unsupervised,sourati2023accelerating,rzhetsky2015choosing,perozzi2014deepwalk,grover2016node2vec}, but applies it as a forecast-time consistency check. For a target such as thermoelectricity, the query is the thermoelectricity concept and the known positives are materials already known to be thermoelectric. Using the trained embedding, we score and sort the candidate materials twice: once by their closeness to the query, and once by their closeness to those known positives, giving two ranked lists over the same candidates. When the two lists place largely different materials near the top, they point to disjoint candidate regions, and the forecast may reflect an unstable or undersampled representation rather than a robust target-relevant signal. When the top of both lists instead contains overlapping materials, the forecast is supported from two directions: the same materials rank highly both because they resemble the target concept and because they resemble materials already known to have it. Agreement from both views is stronger evidence than agreement from either one alone.

QAA serves one primary role and supports two downstream analyses. Its primary role is forecast-time saturation diagnosis across walk budgets; the two downstream analyses use the same query--anchor signal for candidate-level support (Section~\ref{sec:support}) and for post-selection reranking (Section~\ref{sec:reranking}). Figure~\ref{fig:qaa_workflow} summarizes the full workflow.

\begin{figure*}[!t]
    \centering
    \includegraphics[width=\textwidth,keepaspectratio,trim=0in .3in 0in 0in,clip]{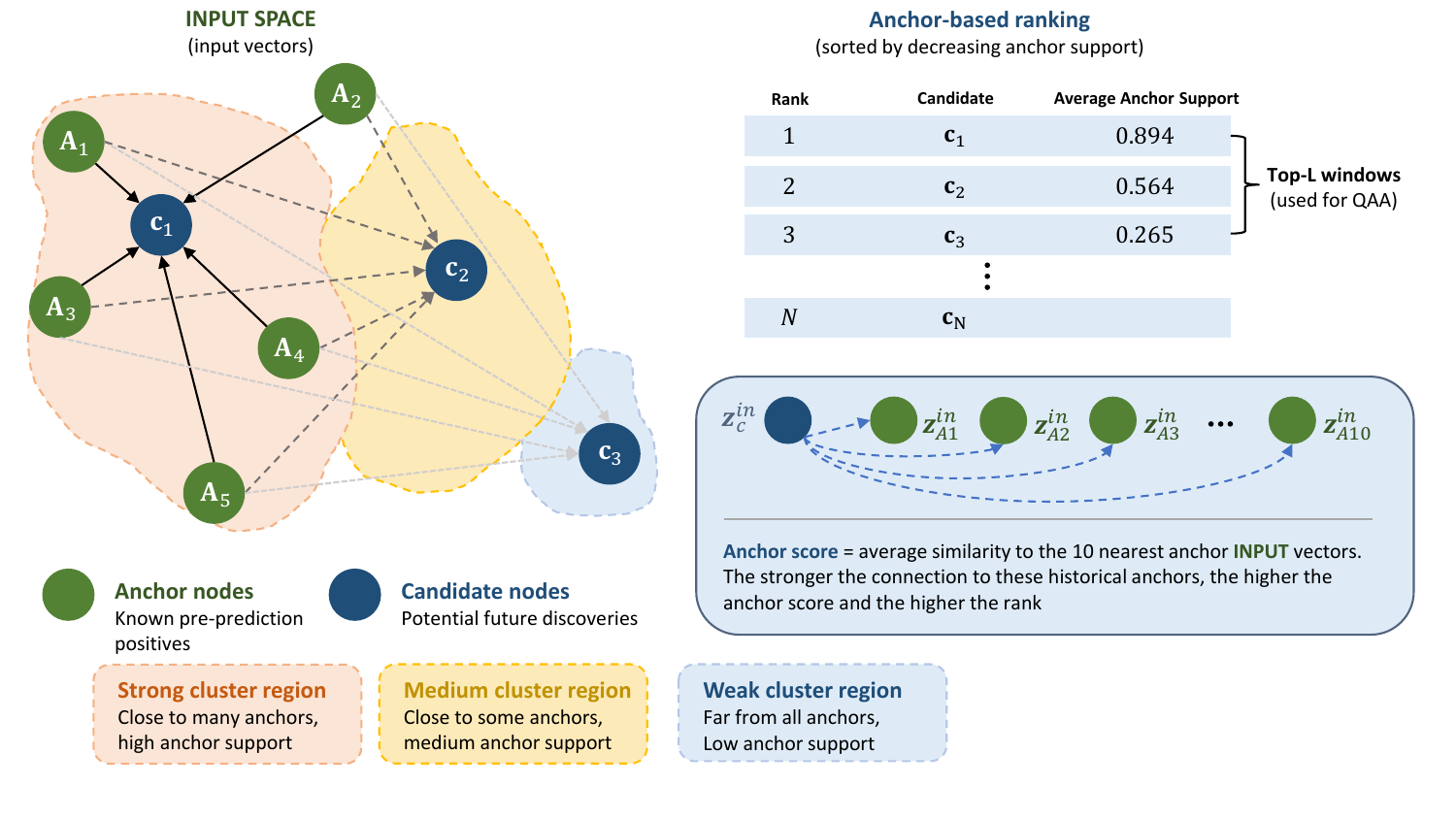}
    \caption{Anchor-induced candidate ranking. Known pre-prediction positives define anchor neighborhoods in the learned input-vector space. Each candidate receives an anchor score from its average similarity to the nearest in-vocabulary anchors, producing a second candidate ranking that is compared with the query-based ranking by QAA.}
    \label{fig:anchor_ranking}
\end{figure*}

\subsection{Query-Based Forecast Ranking}
\label{sec:query_ranking}
The primary output of each walk-based model is a ranked list of candidate entities for a target query. This follows the original benchmark setting: after embeddings are trained from the pre-prediction hypergraph, the target concept is used as a query, and candidates are ranked by their embedding similarity to that query. For thermoelectric forecasting, the query is the thermoelectric property and the ranked candidates are materials. For COVID-19 drug repurposing, the query is the target biomedical concept and the ranked candidates are drugs. This query-based ranking is the model's direct forecast: candidates closer to the query in the learned embedding space are treated as more likely future discoveries. Figure~\ref{fig:query_ranking} illustrates this query-based ranking step.

For each walk policy \(m\), walk budget \(w\), and random seed \(r\), let \(q\) denote the target query and \(c \in \mathcal{C}\) denote a candidate entity. Word2Vec learns two vectors for each node: an input vector, used when the node appears as the center token, and an output vector, used when the node appears as a context token. Following the original benchmark scoring convention, we score query--candidate pairs with input--output similarity because the target query and candidate entities are different node types and the forecast asks how strongly a candidate appears in the context of the query. The query-based forecast score is

\begin{equation}
s_{q,r}^{(m,w)}(c)
=
\cos\left(
z_{q,r}^{(m,w,\mathrm{in})},
z_{c,r}^{(m,w,\mathrm{out})}
\right),
\end{equation}

where \(z_{q,r}^{(m,w,\mathrm{in})}\) is the input vector for the target query and \(z_{c,r}^{(m,w,\mathrm{out})}\) is the output vector for the candidate. Candidates are ranked in descending order of \(s_{q,r}^{(m,w)}(c)\), producing the query-based forecast ranking \(Q_{m,w,r}\). Candidates absent from the embedding vocabulary are retained in the pool but assigned floor scores and worst ranks.

This ranking provides the query-centered view used by QAA. Subsequent sections construct an anchor-induced ranking over the same candidate pool and compare the two rankings to measure whether the target query and its known pre-prediction positives emphasize similar candidate regions.

\subsection{Anchor Construction}
\label{sec:anchor_construction}

We build the anchor set \(\mathcal{A}_q\) for each query \(q\) from the released benchmark's pre-prediction positive labels, keeping only entities dated before the prediction date (materials known to be thermoelectric before 2001; drugs associated with the \emph{coronavirus} property before 2020). For each walk configuration, we further restrict to anchors present in the learned embedding vocabulary, denoted \(\mathcal{A}_{q,r}^{(m,w)}\), since stochastic walk generation and embedding training may exclude some low-frequency nodes.

The COVID-19 task illustrates that the anchoring target need not be the forecast target itself. Because no COVID-19-specific literature exists before 2020, we anchor on the adjacent \emph{coronavirus} property, whose pre-2020 positives are available at forecast time; this aligns the anchor neighborhood with the intended target when the target itself has no pre-prediction history.

\subsection{Anchor-Induced Candidate Ranking}
\label{sec:anchor_ranking}

Using the same candidate pool, we score each candidate by its similarity to the anchor set rather than to the query. Because anchors and candidates are entities of the same type, we use input--input cosine similarity, which treats them as points in the same learned entity space rather than asking whether one appears in the context of the other. Rather than comparing each candidate to all anchors, we score it by its average similarity to the ten nearest in-vocabulary anchors. This local averaging treats anchors as a target-relevant neighborhood: it reduces dependence on any single historical positive while avoiding dilution from anchors that lie in distant parts of the embedding space. We fixed the neighborhood size to \(k_A=10\) before running any experiments and kept it locked across both benchmarks, all policies, and all checkpoints; it was not tuned on results. Figure~\ref{fig:anchor_ranking} illustrates this anchor-induced ranking step.

Let \(k_A=10\), and let \(\mathrm{NN}_{k_A}(c;\mathcal{A}_{q,r}^{(m,w)})\) denote the set of the \(k_A\) nearest in-vocabulary anchors to candidate \(c\) under input--input cosine similarity, or all available in-vocabulary anchors if fewer than \(k_A\) are present. We define the anchor-induced score as

\begin{equation}
s_{A,r}^{(m,w)}(c)
=
\frac{1}{|\mathrm{NN}_{k_A}(c;\mathcal{A}_{q,r}^{(m,w)})|}
\sum_{a \in \mathrm{NN}_{k_A}(c;\mathcal{A}_{q,r}^{(m,w)})}
\cos\left(
z_{c,r}^{(m,w,\mathrm{in})},
z_{a,r}^{(m,w,\mathrm{in})}
\right).
\end{equation}

Candidates with higher average similarity to nearby anchors are ranked higher, producing the anchor-induced ranking \(A_{m,w,r}\). Candidates absent from the embedding vocabulary are retained in the candidate pool but assigned floor scores, so they remain in the full ranking and appear near the bottom.

QAA uses this ranking as a forecast-time comparison point: high top-window overlap between the query-based and anchor-induced rankings indicates that the forecast is supported by both the target query and known pre-prediction examples of the target concept.

\subsection{Multiscale Query--Anchor Agreement}
\label{sec:multiscale_qaa}

The anchor-induced ranking lets us see how strongly each candidate is supported by its proximity to the target's historical examples, and comparing it with the query-based ranking tells us how much the two views agree. To compare them we look at their top candidates, a window, and count how many appear in both. How much they agree depends on how large a window we choose. A window that is too small can make the overlap swing on a few rank changes that amount to noise, while a window that is too large covers so much of the candidate pool that overlap loses meaning. We therefore measure agreement not at a single window but across a series of increasingly broad ones. For each walk policy \(m\), walk budget \(w\), and random seed \(r\), let \(Q_{m,w,r}\) denote the query-based forecast ranking and \(A_{m,w,r}\) denote the anchor-induced ranking. For a candidate window of size \(L\), we define

\begin{equation}
\mathrm{QAA}_{m,w,r,L}
=
\frac{
\left|
\mathrm{Top}_L(Q_{m,w,r})
\cap
\mathrm{Top}_L(A_{m,w,r})
\right|
}{L}.
\end{equation}

Here, \(\mathrm{Top}_L(\cdot)\) denotes the set of the top \(L\) candidates in a ranking. QAA ranges from 0 to 1, with higher values indicating greater top-window overlap between the candidates emphasized by the target query and those emphasized by its pre-prediction anchors.

We compute QAA over multiple candidate-window sizes rather than a single cutoff. The goal is to test whether query--anchor agreement holds across a meaningful range of candidate neighborhoods, not only at one arbitrary ranking depth. Averaging top-window overlap across a geometric series of cutoffs is closely related in spirit to rank-biased overlap, which weights agreement more heavily near the top of two rankings and remains well defined for indefinite or partially overlapping rankings~\citep{webber2010similarity}. Very small windows can be noisy because a few rank swaps or seed-specific fluctuations can substantially change the overlap. Very large windows can also be uninformative because they cover so much of the candidate pool that agreement begins to reflect broad background overlap rather than shared prioritization. We therefore use four broad windows for each benchmark, beginning with a benchmark-specific minimum window large enough to reduce small-set noise and then doubling across successive scales. If the next doubled window would exceed approximately half of the candidate pool, we cap the final window near the half-pool scale. This design captures agreement across increasingly broad candidate regions while keeping the comparison focused on prioritized candidates.

The candidate-window sets used in each benchmark are shown in Table~\ref{tab:qaa_windows}, where the pool size \(N\) is the candidate pool reported in Table~\ref{tab:benchmark_stats}.

\begin{table}[h]
\centering
\small
\begin{tabular}{lcc}
\hline
Benchmark & Candidate pool size \(N\) & \(W(N)\) \\
\hline
Thermo & \(12{,}619\) & \(\{500,1000,2000,4000\}\) \\
COVID & \(3{,}118\) & \(\{250,500,1000,1500\}\) \\
\hline
\end{tabular}
\caption{Candidate-window sizes used for multiscale QAA.}
\label{tab:qaa_windows}
\end{table}

The final multiscale QAA score is the average top-window overlap across these windows:

\begin{equation}
\mathrm{QAA}_{m,w,r}
=
\frac{1}{|W(N)|}
\sum_{L \in W(N)}
\mathrm{QAA}_{m,w,r,L}.
\end{equation}

The diagnostic value of QAA comes from the contrast between agreement and disagreement. When QAA is low, the target query and its known positives induce different candidate neighborhoods: the query-based forecast suggests one region of the embedding space, while the anchor-induced ranking suggests another. In this case, the representation has not organized the target concept, historical positives, and nearby candidates into overlapping candidate regions. We call the representation \emph{target-coherent} when these two views instead retrieve overlapping regions of the candidate pool, that is, when QAA is high. When QAA is low, the resulting forecast may still contain useful candidates, but it lacks internal support from the target's known positive structure.

Target-coherence is useful because it lets us interrogate the forecast before future discovery labels exist: QAA does not prove that the candidates are correct, but it tests whether the forecast is internally supported by the target structure available at prediction time.

We use this agreement as a saturation signal across walk budgets. At low walk counts, random-walk sampling may provide an incomplete and noisy view of the hypergraph, so the query and anchor neighborhoods can shift substantially across configurations. As additional walks are sampled, recurring target-relevant paths are observed more consistently. If QAA rises and then plateaus, the query--anchor geometry has stopped changing substantially, suggesting that the walk process has reached a stable target-relevant regime. Thus, QAA is not a pointwise precision oracle or a substitute for retrospective evaluation; it is a forecast-time diagnostic for whether the representation supporting the forecast is target-coherent and saturated with respect to known pre-prediction positives.

\subsection{QAA Smoothing and Forecast-Time Saturation}
\label{sec:qaa_saturation}

Raw QAA values can fluctuate across neighboring walk budgets because each budget produces a separately trained stochastic embedding. These fluctuations may come from random-walk sampling, embedding initialization, and local instability in the learned neighborhood structure. Since our goal is to identify the onset of saturation rather than the single budget with the highest raw QAA, we smooth the QAA curve before selecting a walk count.

For each walk policy \(m\) and walk budget \(w\), we first aggregate QAA across repeated random seeds:

\begin{equation}
\overline{\mathrm{QAA}}_{m,w}
=
\frac{1}{R}
\sum_{r=1}^{R}
\mathrm{QAA}_{m,w,r},
\end{equation}

where \(R\) is the number of repeated runs for that configuration. This averaging is used to reduce dependence on any single random-walk realization or embedding initialization. The value of \(R\) is part of the experimental protocol rather than a property of QAA itself.

We then estimate the local saturation trend with a centered five-point rolling mean over adjacent walk budgets:

\begin{equation}
\widetilde{\mathrm{QAA}}_{m}(w_i)
=
\frac{1}{|\mathcal{N}_i|}
\sum_{j \in \mathcal{N}_i}
\overline{\mathrm{QAA}}_{m,w_j},
\end{equation}

where \(w_i\) is the \(i\)-th walk budget and \(\mathcal{N}_i=\{i-2,i-1,i,i+1,i+2\}\), truncated at the boundaries of the sweep. In our experiments, walk budgets are spaced in increments of 5{,}000 walks, so this smoother averages over a small local neighborhood of adjacent budgets. The five-point window is not meant to be a universal constant; it is a modest local smoother chosen to damp seed- and budget-specific fluctuations without erasing the broader walk-count trend. For a different sweep resolution, the smoothing window should be chosen relative to the spacing of evaluated budgets.

For each walk policy, let

\begin{equation}
M_m
=
\max_i
\widetilde{\mathrm{QAA}}_{m}(w_i)
\end{equation}

denote the best observed value of the smoothed QAA curve. We define the QAA plateau budget as the earliest walk budget whose smoothed QAA reaches at least \(99\%\) of this value:

\begin{equation}
\hat{w}_m
=
\min
\left\{
w_i :
\widetilde{\mathrm{QAA}}_{m}(w_i)
\geq
0.99 M_m
\right\}.
\end{equation}

This rule selects the first point at which query--anchor agreement has essentially reached its best observed level. The purpose is not to maximize QAA post hoc, but to identify the earliest walk-count regime where additional random walks no longer substantially change the target-relevant geometry of the representation. A single high QAA value indicates internal agreement for one embedding configuration; reaching \(99\%\) of the best observed smoothed QAA indicates that further sampling yields little additional query--anchor agreement across neighboring walk budgets.

We note a practical caveat to this definition. Because \(M_m\) is the maximum smoothed QAA over the evaluated checkpoints, computing it requires observing the sweep up to that point, and our 2.5-million-usable-walk range is itself a bounded window rather than an exhaustive sweep. This does not, however, make the criterion arbitrary. Empirically, smoothed QAA rises and then declines with no sign of subsequent recovery (Figures~\ref{fig:qaa_super_panel} and~\ref{fig:qaa_validation_supergraph}), and because QAA measures saturation of the query--anchor geometry, a trajectory that saturates, declines substantially, and then climbs back to a materially higher level is unlikely. The observed maximum is therefore a reasonable reference for the plateau rule within the evaluated range. In practice, the criterion admits two usable forms. An online form watches for a plateau followed by a sustained decline with no recovery and stops there, while a bounded-sweep form evaluates a fixed budget window and selects the plateau afterward. The need for such a criterion does not disappear even with a completed sweep. Because additional walks are not monotonically beneficial, the final checkpoint is generally not the best one, so a signal like QAA is still required to choose where to stop rather than simply sampling as long as possible.

The plateau rule is applied separately within each walk policy. QAA therefore selects a saturated walk-count regime conditional on the chosen walk policy; it does not choose among walk policies. Post-prediction discovery labels are not used to compute QAA, smooth the QAA curve, or select \(\hat{w}_m\). Those labels are reserved for retrospective evaluation of the QAA-selected budget.

\subsection{Candidate-Level Query--Anchor Support}
\label{sec:support}
The multiscale QAA score shows how much the query-based and anchor-induced rankings agree overall, for one configuration of walk policy, budget, and seed. Here we look into the region where that agreement is strongest, the candidates ranked highly by both the query and the anchors, and ask whether these jointly supported candidates are the ones most likely to become future discoveries. This tests whether query--anchor agreement identifies internally supported candidates, those backed by both the target and its historical examples, and not just whether a walk-count regime has saturated overall.

For each candidate \(c\), let \(r_q^{(m,w,r)}(c)\) denote its rank in the query-based forecast ranking and \(r_A^{(m,w,r)}(c)\) denote its rank in the anchor-induced ranking. Candidate-level support is defined over the saved analysis pool for each dataset, checkpoint, and seed. For thermoelectric forecasting, this pool contains the saved top-5{,}000 candidates; for COVID-19, where the candidate pool is smaller, it contains the full candidate pool. Let \(N_{\mathrm{pool}}\) denote the size of this analysis pool.

For a relative support threshold \(\rho\), we define the rank cutoff

\begin{equation}
\tau_{\rho}=\left\lceil \rho N_{\mathrm{pool}} \right\rceil .
\end{equation}

We evaluate \(\rho\) from \(0.01\) to \(0.50\) in steps of \(0.01\), corresponding to support definitions ranging from strict top-region agreement to broad top-half agreement. For each threshold, candidates are assigned to four groups:

\begin{equation}
\begin{aligned}
\text{Query+Anchor}_{\rho} &= \{c : r_q(c) \leq \tau_{\rho},\ r_A(c) \leq \tau_{\rho}\}, \\
\text{Query-Only}_{\rho} &= \{c : r_q(c) \leq \tau_{\rho},\ r_A(c) > \tau_{\rho}\}, \\
\text{Anchor-Only}_{\rho} &= \{c : r_q(c) > \tau_{\rho},\ r_A(c) \leq \tau_{\rho}\}, \\
\text{Neither}_{\rho} &= \{c : r_q(c) > \tau_{\rho},\ r_A(c) > \tau_{\rho}\}.
\end{aligned}
\end{equation}

These groups are formed using only forecast-time rankings at the QAA-selected plateau checkpoints. Post-prediction discovery labels are used only afterward to evaluate future-discovery rate, enrichment over the analysis pool, and cumulative precision within each support group. The purpose of this analysis is to test whether query--anchor agreement provides candidate-level support beyond the direct query ranking alone.

\subsection{Anchor-Supported Reranking}
\label{sec:reranking}
After QAA identifies a saturated walk regime, we test whether the anchor-induced ranking can also be used to refine candidate ordering. This reranking step is a post-selection analysis rather than part of the QAA saturation criterion. The goal is to use the anchor-induced ranking to reorganize the query-based ranking, pushing up candidates supported by both and suppressing those supported by only one.

We evaluate two variants that differ in how strongly the anchor-induced ranking is allowed to affect the final forecast. The uncapped variant gives the anchor signal full influence alongside the query-based ranking, allowing anchors to substantially reorder candidates. This makes uncapped reranking a high-leverage exploratory variant: it can recover candidates missed by the direct query ranking, but it can also over-promote candidates that are close to historical positives for reasons that do not improve future discovery recovery. The capped variant is more conservative. It uses the anchor ranking only as support for candidates that are also supported by the query-based ranking, so anchors can boost but not dominate the direct forecast.

At the QAA-selected walk budget \(\hat{w}_m\), we combine the query-based and anchor-induced rankings using reciprocal rank fusion (RRF)~\citep{cormack2009reciprocal}. RRF converts ranks into reciprocal-rank support scores, allowing rankings with different score scales to be combined without directly comparing their raw similarity values. The ranks \(r_q\) and \(r_A\) are taken over the same configured candidate pool used for the query-based and anchor-induced rankings, with out-of-vocabulary candidates held at their floor (worst) ranks. Following the standard RRF formulation, we use an offset of 60, which prevents the fused score from being dominated only by the top few ranks and makes support decay more gradually with rank. To avoid overloading the ranking symbols \(Q_{m,w,r}\) and \(A_{m,w,r}\), we write the query-ranking reciprocal-rank contribution for seed \(r\) as

\begin{equation}
g_{q,r}^{(m)}(c)
=
\frac{1}{60 + r_q^{(m,\hat{w}_m,r)}(c)},
\end{equation}

and the anchor-ranking contribution is

\begin{equation}
g_{A,r}^{(m)}(c)
=
\frac{1}{60 + r_A^{(m,\hat{w}_m,r)}(c)}.
\end{equation}

Here, \(r_q^{(m,\hat{w}_m,r)}(c)\) is the rank of \(c\) in the query-based forecast ranking, and \(r_A^{(m,\hat{w}_m,r)}(c)\) is its rank in the anchor-induced ranking. The constant 60 follows the reciprocal-rank fusion offset used by \citet{cormack2009reciprocal}; it smooths rank contributions so that scores decay gradually with rank rather than being dominated only by the top few positions.

We evaluate two variants. The first is an uncapped anchor-supported reranking score:

\begin{equation}
S_{\mathrm{uncapped},r}^{(m)}(c)
=
g_{q,r}^{(m)}(c)
+
g_{A,r}^{(m)}(c).
\end{equation}

This variant gives the anchor-induced ranking full influence alongside the query-based ranking, corresponding to the exploratory, higher-variance option described above.

The second variant is a capped anchor-supported reranking score:

\begin{equation}
S_{\mathrm{capped},r}^{(m)}(c)
=
g_{q,r}^{(m)}(c)
+
\min\left(g_{A,r}^{(m)}(c), g_{q,r}^{(m)}(c)\right).
\end{equation}

The first term preserves the original query-based forecast, while the second term adds anchor support only up to the strength of the query support. Thus, capped reranking constrains anchors to act as supporting evidence rather than as an independent replacement for the query signal, the more conservative of the two options introduced above.

\begin{table*}[!t]
\centering
\small
\resizebox{\linewidth}{!}{%
\begin{tabular}{lcc}
\hline
Statistic & Thermoelectric materials & COVID-19 therapeutics \\
\hline
Target query & Thermoelectricity & COVID-19 therapeutic relevance \\
Candidate type & Materials & Drugs \\
Training window & 1996--2000 & 2015--2019 \\
Prediction date & 2001 & Jan. 1, 2020 \\
Evaluation horizon & 2001--2018 & Jan. 2020--July 2021 \\
Evaluation cadence & Annual & Monthly \\
\hline
Training records & 85{,}517 & 4{,}271{,}479 \\
Historical target-linked candidates & 362 & 129 \\
Candidate pool size & 12{,}619 & 3{,}118 \\
\hline
All future target positives & 3{,}720 & 279 \\
Future positives in candidate pool & 828 & 279 \\
Candidate-pool positive rate & 6.56\% & 8.95\% \\
\hline
\end{tabular}%
}
\caption{Benchmark statistics for the two prospective discovery forecasting tasks. Historical target-linked candidates are candidate-type entities associated with the target before the prediction date. All future target positives are entities that become associated with the target during the evaluation horizon; future positives in the candidate pool are the subset already observed in the training data and scored by the model. Evaluation focuses on this candidate-pool subset because positives outside the pool, such as newly appearing thermoelectric materials, are absent from the pre-prediction graph and therefore outside the benchmark's scoring universe. For COVID-19, the 129 historical target-linked candidates are the pre-2020 coronavirus-associated drugs used as QAA anchors, whereas the 279 future positives are the distinct drugs entering COVID-19 clinical trials during the evaluation horizon; the two quantities measure different things.}
\label{tab:benchmark_stats}
\end{table*}

%% file: sections/experimental_setup.tex
\subsection{Prediction Tasks and Benchmark Datasets}

We evaluate QAA on two temporally structured prospective discovery benchmarks released by \citet{sourati2023accelerating}: thermoelectric materials discovery and COVID-19 drug repurposing. In both tasks, the model is trained only on information available before a fixed prediction date, ranks a predefined candidate pool, and is evaluated against discoveries observed after that date. We preserve the original benchmark definitions, including the training windows, candidate pools, prediction dates, and post-prediction discovery labels. Table~\ref{tab:benchmark_stats} summarizes the main task statistics.

\subsubsection{Thermoelectric Materials Discovery}

Following \citet{sourati2023accelerating}, we evaluate prospective forecasting for thermoelectric materials using their released preprocessed benchmark. The benchmark is derived from a Scopus-indexed inorganic materials corpus spanning 1937--2019. We do not reconstruct the corpus from raw Scopus records; all dataset statistics reported here are computed from the released benchmark files and summarized in Table~\ref{tab:benchmark_stats}.

The prediction task asks which candidate materials will first become associated with thermoelectricity after the 2001 prediction date. The training hypergraph is constructed from the fixed five-year publication window 1996--2000, matching the original benchmark setting. This window contains \textbf{85{,}517} publication records.

The candidate pool consists of materials mentioned during the training window that had not been associated with thermoelectricity before 2001. After graph-connectivity filtering, which requires each candidate to have at least one incident hyperedge in the training hypergraph, \textbf{12{,}619} candidate materials remain for evaluation. During the 2001--2018 evaluation horizon, \textbf{3{,}720} materials become associated with thermoelectricity, of which \textbf{828} are present in the prediction-time candidate pool and can therefore receive model scores under the benchmark protocol.

Ground-truth discoveries are taken from the Sourati and Evans release, which follows the benchmark introduced by \citet{tshitoyan2019unsupervised}. In this setting, first-time thermoelectric associations are identified by co-occurrence between material formulas and the thermoelectric property keyword in publication abstracts. We evaluate thermoelectric forecasts using annual cumulative precision from 2001 through 2018, focusing on future positives in the candidate pool.

\subsubsection{COVID-19 Drug Repurposing}

We evaluate prospective COVID-19 drug repurposing using the preprocessed biomedical benchmark released by \citet{sourati2023accelerating}, derived from MEDLINE, DrugBank, and ClinicalTrials.gov. The prediction date is January 1, 2020, immediately before the emergence of COVID-19-related biomedical literature and clinical trial activity. The training hypergraph uses publications from 2015--2019 and contains \textbf{4{,}271{,}479} records.

The candidate pool is the DrugBank-derived drug vocabulary released with the benchmark, restricted to drugs that appear in the 2015--2019 training hypergraph. After this intersection, the fixed evaluation pool contains \textbf{3{,}118} candidate drugs. Unlike the thermoelectric task, all \textbf{279} future COVID-19 therapeutic positives in this benchmark are contained in the candidate pool, so the all-positive and candidate-pool-positive counts coincide.

Ground-truth discoveries are drugs that entered COVID-19-related clinical trials between January 2020 and July 2021, following \citet{gysi2021network}. We evaluate COVID-19 forecasts using monthly cumulative precision over this evaluation horizon.

\subsection{Hypergraph Construction and Random-Walk Implementation}

We instantiate the human-aware discovery setting of \citet{sourati2023accelerating} using their released benchmark artifacts and candidate lists, while preserving their post-prediction discovery labels only for retrospective evaluation. We preserve the original mixed-hypergraph representation described in Section~\ref{sec:background}, in which each publication is a hyperedge over typed author, material or drug, and property or concept nodes. This preserves author-mediated pathways, where materials, drugs, and target concepts can be connected indirectly through scientists whose publication histories span otherwise weakly connected domains.

Our one departure from the original benchmark's sampling procedure is in how the walk budget is counted: we count only the valid walks that survive preprocessing, rather than every raw generated walk, so that a walk-budget checkpoint corresponds to a fixed amount of usable training data across policies (Section~\ref{sec:policies_checkpoints}).

\subsection{Random-Walk Policies and Usable-Corpus Checkpoints}
\label{sec:policies_checkpoints}

We evaluate the three walk policies defined in Section~\ref{sec:background} (uniform, \(\alpha=1\), and \(\alpha\to\infty\)), preserving the original benchmark's traversal semantics.

A key difference from a raw walk-count sweep is that our checkpoints are defined by \emph{usable Word2Vec training lines}. Raw generated walks can become unusable after preprocessing because author tokens are removed and some walks collapse to empty, degenerate, or too-short sequences. As a result, the same nominal number of raw walks can yield different amounts of actual Word2Vec training data across policies. Table~\ref{tab:usable_retention} reports this fixed-raw diagnostic: after generating 2.5 million raw walks, the number of usable training lines differs substantially by traversal policy. In the COVID-19 diagnostic, for example, 2.5 million raw walks produce only 1.44 million usable lines under uniform traversal, compared with 2.08 million under alpha-1 and 2.5 million under alpha-inf. A raw-walk checkpoint would therefore compare policies trained on unequal effective corpus sizes.

\begin{table}[t]
\centering
\small
\begin{tabular}{llrrr}
\hline
Benchmark & Policy & Raw walks & Usable lines & Retention \\
\hline
Thermo & Uniform & 2{,}500{,}000 & 2{,}213{,}107 & 88.5\% \\
Thermo & \(\alpha=1\) & 2{,}500{,}000 & 2{,}367{,}437 & 94.7\% \\
Thermo & \(\alpha\to\infty\) & 2{,}500{,}000 & 2{,}500{,}000 & 100.0\% \\
\hline
COVID & Uniform & 2{,}500{,}000 & 1{,}441{,}038 & 57.6\% \\
COVID & \(\alpha=1\) & 2{,}500{,}000 & 2{,}078{,}896 & 83.2\% \\
COVID & \(\alpha\to\infty\) & 2{,}500{,}000 & 2{,}500{,}000 & 100.0\% \\
\hline
\end{tabular}
\caption{Usable-walk retention under fixed raw-walk generation. Retention is the percentage of raw generated walks that remain as usable no-author Word2Vec training lines after preprocessing. Thermoelectric and COVID-19 values are averages over ten seeds.}
\label{tab:usable_retention}
\end{table}

To remove this confound, each policy--seed run generates a persisted valid walk corpus. Walk generation continues until the required number of usable no-author training lines is obtained, and checkpoints are then defined as prefixes of this usable corpus. Thus, a checkpoint of 250{,}000 denotes 250{,}000 valid Word2Vec training sequences for every walk policy, rather than 250{,}000 raw generated walks. For each checkpoint, we train a new Word2Vec model from scratch. Checkpoints increase in increments of 5{,}000 usable lines, and each of the ten random seeds has its own independently generated walk corpus. Reported results average over these seeds. Unless otherwise noted, each reported metric is the mean over the ten seeds, and error terms are 95\% confidence intervals given as the mean plus or minus a \(t\)-based half-width (\(t\)-distribution, \(\mathrm{df}=9\)). For reranking comparisons, the direct and reranked scores are not seed-paired in the source aggregates, so where their intervals are reported they use a conservative independent-standard-error approximation.

\subsection{Embedding Training}

We train a separate Word2Vec model from scratch at each usable-corpus checkpoint for each dataset, walk policy, and random seed. Thus, each checkpoint corresponds to an independently trained embedding model using a prefix of the persisted usable walk corpus for that policy--seed run. The checkpoint sweep therefore measures how forecast quality and QAA change as the amount of usable random-walk context increases, rather than continuing training from a previous checkpoint.

The learned Word2Vec model provides input and output vectors for each node. As described in Section~\ref{sec:qaa}, query-based forecasts use input--output cosine similarity between the target query and candidate entities, while anchor-induced rankings use input--input similarity between candidates and pre-prediction anchors. Unless otherwise stated, Word2Vec hyperparameters follow the released benchmark implementation of \citet{sourati2023accelerating}. All models are trained with gensim~\citep{rehurek2010gensim,gensim_software}.

\subsection{Baselines}

We include the transition-based baselines from \citet{sourati2023accelerating}, which we denote S2 and S3 (corresponding to their two- and three-step transition metrics, \(s=2\) and \(s=3\)), along with a random-ranking baseline. S2 and S3 are non-embedding graph baselines that score candidates by transition structure in the pre-prediction hypergraph. S2 ranks candidates by the strength of short two-step transitions from the target query to candidate entities, while S3 extends this idea to three-step transitions, allowing one additional intermediate hop through the graph. These baselines therefore test how much forecast signal can be recovered from local graph connectivity without learning Word2Vec embeddings.

The random baseline ranks candidates uniformly at random and provides a chance-level reference for each candidate pool. We use the released baseline scores under the original benchmark protocol rather than tuning these baselines over the walk-count sweep.

These baselines are fixed for each benchmark and do not vary across usable-corpus checkpoints, embedding seeds, or QAA plateau choices. They are therefore used only as reference points for contextualizing the performance of the walk-based embedding forecasts. Because QAA requires both a query-based embedding ranking and an anchor-induced embedding ranking, QAA-based saturation selection, candidate-level support analysis, and anchor-supported reranking are applied only to walk-based embedding configurations.

\subsection{Evaluation Metrics}
\label{sec:eval_metrics}

All forecasts are produced using only pre-prediction information. Embeddings are trained on the pre-prediction hypergraph, candidate rankings are generated at the prediction date, and QAA-based checkpoint selection is performed without access to post-prediction discovery labels. Post-prediction labels are used only for retrospective evaluation.

\paragraph{Cumulative precision (CP@\(K\))}
We evaluate forecast quality using cumulative precision at \(K\), following the temporal evaluation horizons of the original benchmarks~\citep{sourati2023accelerating}. For a forecast produced at prediction time \(t_0\), cumulative precision is

\begin{equation}
\mathrm{CP@}K(t)
=
\frac{
\left|
\mathrm{Top}_K
\cap
\mathrm{Discoveries}_{[t_0,t]}
\right|
}{K},
\end{equation}

where \(\mathrm{Top}_K\) is the set of top-\(K\) candidates ranked at prediction time and \(\mathrm{Discoveries}_{[t_0,t]}\) is the set of candidates discovered by evaluation time \(t\). We report final CP@50 and CP@250 at the terminal evaluation date for each benchmark. CP@50 measures high-confidence recovery near the top of the ranking, while CP@250 captures broader candidate recovery and is less sensitive to small rank fluctuations near the top of the list.

\paragraph{Future-discovery rate (FDR)}
For candidate-level support analyses, we also report future-discovery rate and enrichment. For a candidate group \(G\), the future-discovery rate is

\begin{equation}
\mathrm{FDR}(G)
=
\frac{
|\mathcal{D}^{+} \cap G|
}{
|G|
},
\end{equation}

where \(\mathcal{D}^{+}\) is the set of post-prediction discoveries in the analysis pool.

\paragraph{Enrichment}
We define enrichment relative to the corresponding analysis pool \(\mathcal{P}\) as

\begin{equation}
\mathrm{Enrichment}(G)
=
\frac{
|\mathcal{D}^{+} \cap G|/|G|
}{
|\mathcal{D}^{+} \cap \mathcal{P}|/|\mathcal{P}|
}.
\end{equation}

Values greater than 1 indicate that a group contains future discoveries at a higher rate than the reference pool. Enrichment is used as a diagnostic of candidate-level support, not as a replacement for rank-based cumulative precision.

\begin{figure*}[!t]
    \centering
    \includegraphics[
        width=\textwidth,
        height=0.78\textheight,
        keepaspectratio
    ]{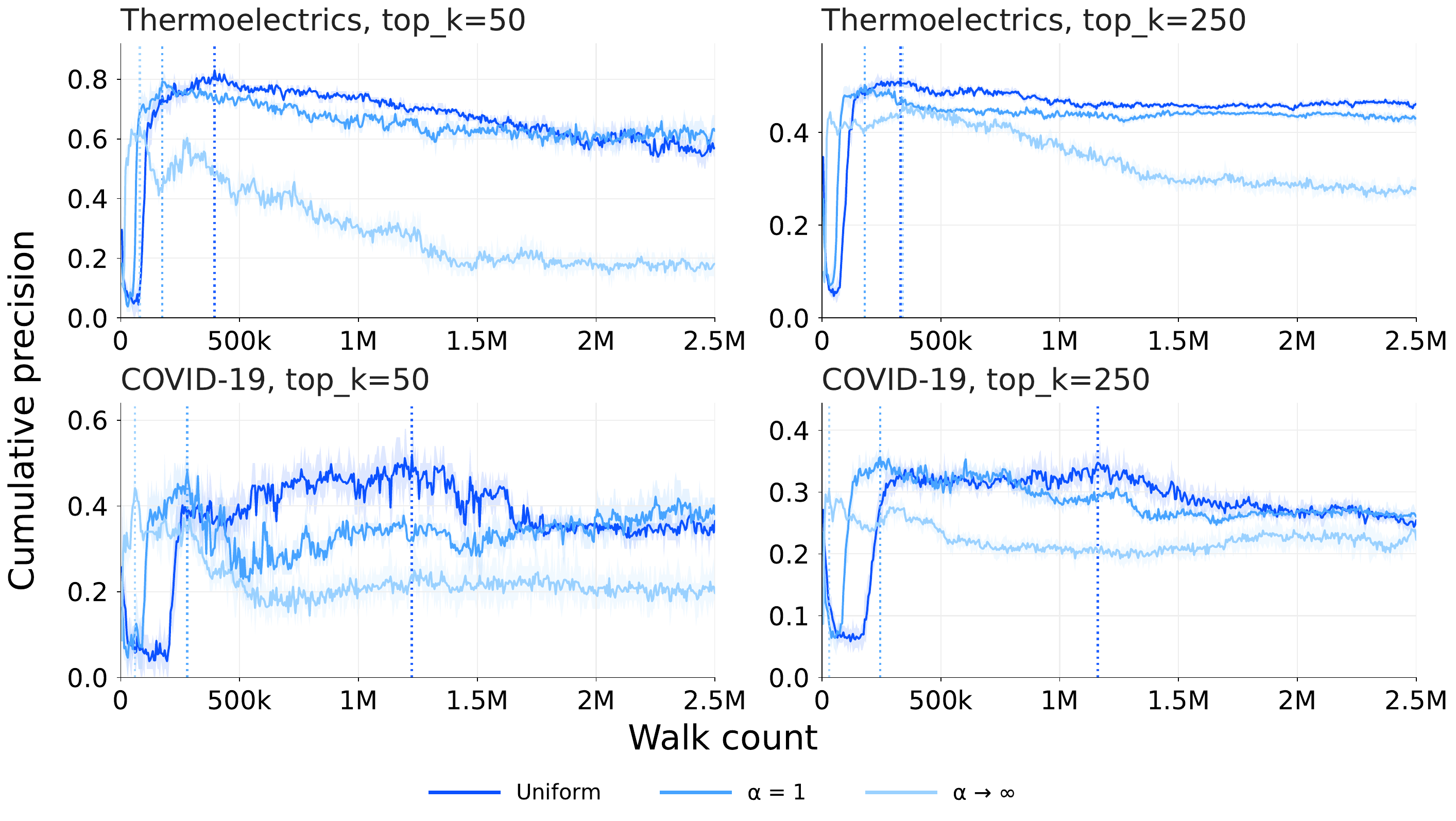}
    \caption{Walk-count sensitivity of cumulative precision across benchmarks, evaluation depths, and walk policies. Each curve shows cumulative precision as a function of usable-corpus checkpoint for one walk policy, averaged over ten seeds; the shaded band shows the seed min--max range. Vertical dotted lines mark each policy's peak checkpoint. Across all twelve settings, precision rises to an interior peak and then declines with additional walks, so neither a fixed nor a uniformly large walk budget is a reliable choice, and the location of the useful region differs by benchmark and policy.}
    \label{fig:walk_sensitivity_panel}
\end{figure*}

\subsection{Retrospective Selection and Diagnostic Comparisons}

To evaluate whether QAA identifies useful walk-count regimes without post-prediction labels, we compare the QAA-selected checkpoint \(\hat{w}_m\) with the retrospectively best checkpoint for each walk policy \(m\) and evaluation depth \(K\). Let

\begin{equation}
w^\star_{m,K}
=
\arg\max_w
\overline{\mathrm{CP@}K}_{m}(w)
\end{equation}

denote the checkpoint with the highest final cumulative precision averaged over random seeds. This retrospective optimum is an oracle-style reference because it uses post-prediction discovery labels and is therefore unavailable at forecast time.

We summarize QAA selection quality using selection regret:

\begin{equation}
\mathrm{SelectionRegret}_{m,K}
=
\max_w
\overline{\mathrm{CP@}K}_{m}(w)
-
\overline{\mathrm{CP@}K}_{m}(\hat{w}_m).
\end{equation}

Selection regret measures the performance cost of using the QAA-selected checkpoint rather than the best checkpoint selected with future labels. We also report the walk-count gap,

\begin{equation}
\mathrm{WalkGap}_{m,K}
=
\left|
\hat{w}_m
-
w^\star_{m,K}
\right|,
\end{equation}

which measures how far the QAA-selected checkpoint is from the retrospective optimum in usable-corpus size. Selection regret is the primary comparison because nearby checkpoints can have similar forecast quality even when their walk-count gap is large.

For candidate-level QAA analyses, support groups are formed before observing post-prediction labels and then evaluated retrospectively using future-discovery rate, enrichment, CP@50, and CP@250. These diagnostic analyses are not used to compute QAA or select \(\hat{w}_m\); they are used to interpret the downstream behavior of QAA-selected regimes.

%% file: sections/results.tex
We first show why a forecast-time diagnostic is needed. Throughout, forecast quality is measured by cumulative precision at depth \(K\) (CP@\(K\)), defined in Section~\ref{sec:eval_metrics}. Across both benchmarks, forecast quality varies substantially with walk policy and usable-corpus checkpoint, and additional walks do not always improve performance: in some cases performance recovers only after a long low-quality region, and in others early gains degrade as more walks are added. We then evaluate QAA as a forecast-time saturation criterion, comparing the checkpoints it selects with retrospective-best checkpoints chosen using post-prediction labels, which we treat as an oracle reference unavailable in prospective use. Finally, we examine whether the same anchor signal can improve candidate ordering after QAA has selected a stable regime, through a candidate-level support analysis and anchor-supported reranking.

\subsection{Walk-Count Sweeps Reveal Configuration Sensitivity}

Figure~\ref{fig:walk_sensitivity_panel} shows why walk-count selection is a serious practical issue in human-mediated discovery forecasting. The curves report cumulative precision across usable-corpus checkpoints for each benchmark, evaluation depth, and walk policy. The number of walks controls how much of the target neighborhood the model sees, so the same policy can look strong or weak depending only on where sampling stops.

At very small checkpoints, this exposure is incomplete. The model has sampled only a limited set of paths, so its view of the target neighborhood can be narrow or unstable. It may strongly emphasize candidates that are close to the first sampled paths while missing other relevant regions of the graph. This explains why several panels begin in a low-performance region and improve only after more usable walk context has been added. Evaluating a walk policy too early can therefore make a potentially useful traversal strategy appear ineffective.

However, the opposite rule is also unsafe: more walks are not always better. Once the walk corpus becomes large, the model may incorporate increasingly broad graph structure that is less specific to the target forecast. Additional walks can mix target-relevant paths with more diffuse background connectivity, weakening the candidate neighborhood that was most useful for prediction. This does not usually appear as a sudden collapse. Instead, several panels show a gradual decline after an earlier high-performing region.

This pattern is clearest in the thermoelectric benchmark. Across the uniform, \(\alpha=1\), and \(\alpha\to\infty\) policies, cumulative precision rises quickly, reaches an early strong region, and then degrades over later checkpoints. The decline is not merely cosmetic: under \(\alpha\to\infty\) at CP@50, thermoelectric precision peaks near \(0.64\) and then falls below \(0.20\) by the end of the sweep, more than halving its best value. The effect appears at both CP@50 and CP@250, although CP@50 is more volatile because small changes near the top of the ranking have a larger effect. The COVID-19 benchmark is noisier, but it shows the same basic issue: performance can recover, plateau, and then deteriorate depending on the policy and checkpoint.

The location of the useful region also changes across settings. A checkpoint that is near a peak for one policy can be too early or too late for another, and the usable-corpus size needed for thermoelectric forecasting does not transfer cleanly to COVID-19. This is expected because the two benchmarks expose different graph structures. The thermoelectric corpus is smaller and more specialized, while the COVID-19 benchmark is much larger, denser, and more heterogeneous, with many drugs, biomedical concepts, and author-mediated paths competing for attention. The same walk policy therefore samples different kinds of neighborhoods in the two tasks.

Walk policy further amplifies this difference. Uniform and \(\alpha=1\) walks preserve more author-mediated exposure, so their useful regimes depend on how author density and publication connectivity shape access to the target neighborhood. In a dense biomedical graph, where a very large publication volume and heavily shared authorship connect many drugs and concepts through common hyperedges, additional walks may continue introducing new paths for much longer, but those paths can also mix target-relevant structure with broad background biomedical connectivity. By contrast, \(\alpha\to\infty\) suppresses much of this author-mediated structure and concentrates more quickly on entity-to-entity neighborhoods, producing earlier but often narrower useful regions.

These differences matter because the walk process is exposing graph structure, not directly observing future scientific truth. A strong checkpoint is therefore one where the sampled paths have provided enough target-relevant structure to support the forecast, but not so much diffuse structure that the target neighborhood becomes diluted. The optimal balance depends on the benchmark, the traversal policy, and the density of the surrounding scientific graph. This is why a single fixed walk budget is not a reliable rule across tasks.
The main implication is that neither a fixed checkpoint nor a uniformly large checkpoint is a reliable prospective rule. Small checkpoints can undersample the target-relevant structure, while very large checkpoints can move the representation away from its strongest forecast region. Since the future labels needed to identify the best checkpoint are unavailable at prediction time, the model needs a forecast-time diagnostic for deciding when the learned representation has stabilized. QAA is designed for this role: it tests whether the query and its known pre-prediction anchors point to overlapping candidate regions before post-prediction discovery outcomes are observed.

\subsection{QAA Identifies Forecast-Time Saturation Regimes}

\begin{table*}[t]
\centering
\small
\setlength{\tabcolsep}{3.8pt}
\renewcommand{\arraystretch}{1.12}
\resizebox{\linewidth}{!}{%
\begin{tabular}{lcl rr cc r}
\hline
Benchmark & \(K\) & Policy & QAA ckpt & Best ckpt & QAA CP (95\% CI) & Best CP (95\% CI) & Regret (pp) \\
\hline
Thermo & 50  & Uniform & 355k & 395k  & 79.00 {\scriptsize $\pm$0.75} & 83.00 {\scriptsize $\pm$1.01} & 4.00 \\
Thermo & 250 & Uniform & 355k & 330k  & 50.52 {\scriptsize $\pm$0.38} & 51.08 {\scriptsize $\pm$0.47} & 0.56 \\
Thermo & 50  & \(\alpha=1\) & 215k & 175k  & 77.00 {\scriptsize $\pm$1.01} & 79.20 {\scriptsize $\pm$0.74} & 2.20 \\
Thermo & 250 & \(\alpha=1\) & 215k & 180k  & 49.00 {\scriptsize $\pm$0.34} & 50.12 {\scriptsize $\pm$0.60} & 1.12 \\
Thermo & 50  & \(\alpha\to\infty\) & 35k & 80k   & 58.00 {\scriptsize $\pm$1.78} & 64.20 {\scriptsize $\pm$0.81} & 6.20 \\
Thermo & 250 & \(\alpha\to\infty\) & 35k & 340k  & 44.40 {\scriptsize $\pm$0.30} & 46.16 {\scriptsize $\pm$0.31} & 1.76 \\
\hline
COVID & 50  & Uniform & 410k & 1225k & 37.40 {\scriptsize $\pm$2.87} & 52.00 {\scriptsize $\pm$2.61} & 14.60 \\
COVID & 250 & Uniform & 410k & 1160k & 31.52 {\scriptsize $\pm$1.09} & 34.76 {\scriptsize $\pm$0.96} & 3.24 \\
COVID & 50  & \(\alpha=1\) & 190k & 280k  & 37.00 {\scriptsize $\pm$2.26} & 48.00 {\scriptsize $\pm$2.52} & 11.00 \\
COVID & 250 & \(\alpha=1\) & 190k & 245k  & 32.92 {\scriptsize $\pm$0.86} & 35.60 {\scriptsize $\pm$0.70} & 2.68 \\
COVID & 50  & \(\alpha\to\infty\) & 25k & 60k   & 33.80 {\scriptsize $\pm$2.07} & 44.20 {\scriptsize $\pm$0.81} & 10.40 \\
COVID & 250 & \(\alpha\to\infty\) & 25k & 30k   & 28.44 {\scriptsize $\pm$0.78} & 30.24 {\scriptsize $\pm$0.62} & 1.80 \\
\hline
\end{tabular}%
}
\caption{QAA-selected checkpoints compared with retrospective-best checkpoints. \(K\) denotes CP@\(K\). Best checkpoints use post-prediction discovery labels and are unavailable at forecast time. Regret is the cumulative-precision gap between the retrospective best and the QAA-selected checkpoint. Reported values are seed means with 95\% confidence intervals (see Section~\ref{sec:experimental_setup}). Overall, QAA-selected checkpoints stay close to the hindsight-best for broad recovery, with CP@250 regret at most 3.24 points across all six benchmark--policy settings; CP@50 regret is larger (up to 14.60 points on COVID-19), reflecting that sharp top-of-ranking selection is harder than broad candidate recovery.}
\label{tab:qaa_selection_quality}
\end{table*}

\begin{figure*}[!t]
    \centering
    \includegraphics[
        width=\textwidth,
        height=0.78\textheight,
        keepaspectratio
    ]{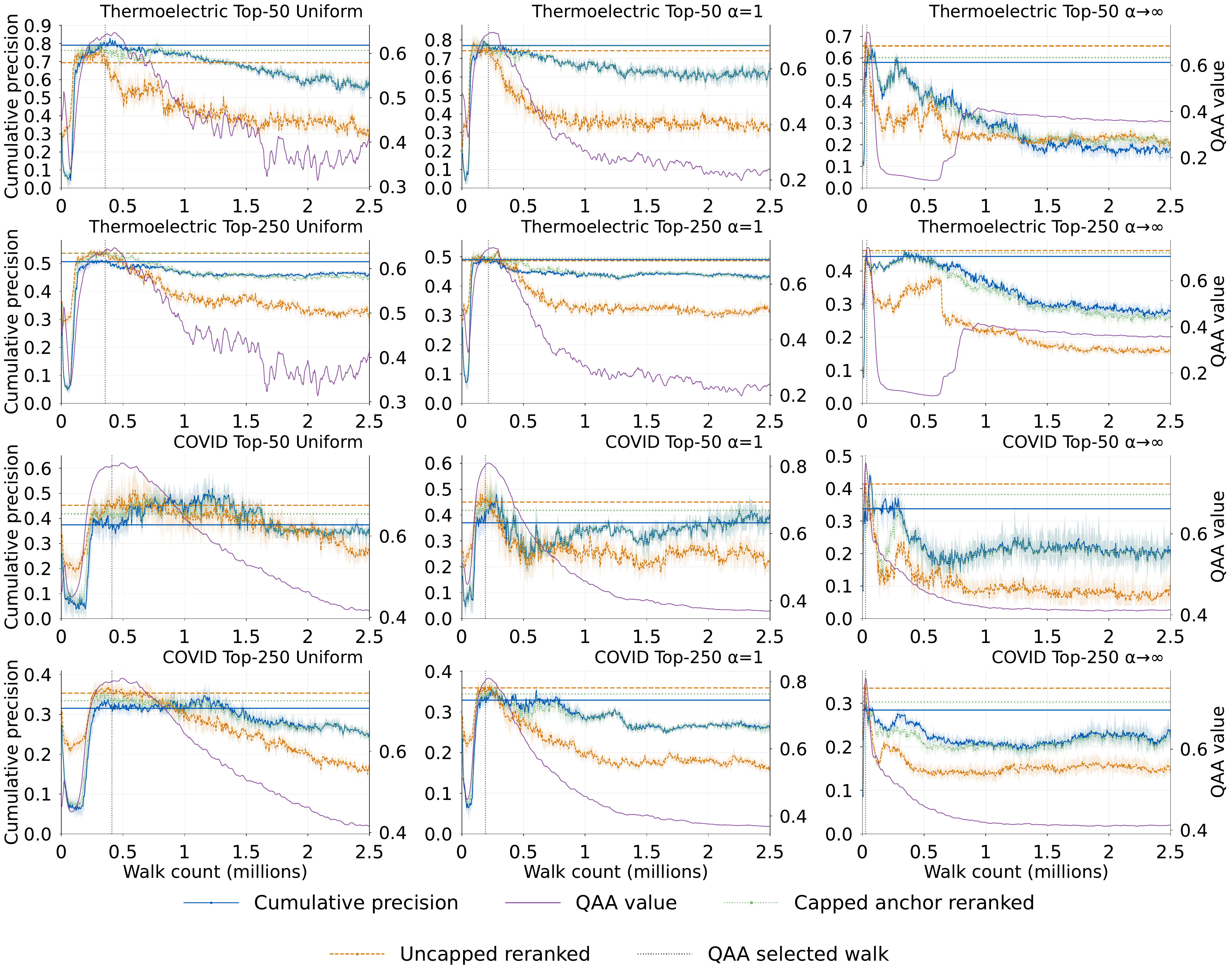}
    \caption{QAA-selected saturation points and anchor-supported reranking across benchmarks, evaluation depths, and walk policies. Blue curves show cumulative precision across usable-corpus checkpoints; purple curves show QAA, computed without post-prediction discovery labels. Vertical dotted lines mark QAA-selected checkpoints. Horizontal reference lines show capped and uncapped anchor-supported reranking at the QAA-selected checkpoint. QAA often selects near a useful saturation region, and reranking can recover or improve performance, especially at Top-250.}
    \label{fig:qaa_super_panel}
\end{figure*}

Figure~\ref{fig:qaa_super_panel} places the QAA signal alongside forecast quality for every benchmark, evaluation depth, and walk policy, so we can see where QAA-based selection lands relative to the precision curves it never observes. Its purple curves show QAA across the same usable-corpus checkpoint sweeps, and the vertical dotted lines mark the checkpoints selected by the smoothed QAA plateau rule, where the smoothed QAA curve first reaches 99\% of its maximum value within the evaluated checkpoint sweep.

Because the future discoveries needed to compute cumulative precision do not exist at prediction time, QAA offers a weaker but available signal: a rising QAA curve means additional walks are still changing the relationship between the query-based and anchor-induced rankings, while a plateau means the two have begun to point to a stable region of candidate space.

Table~\ref{tab:qaa_selection_quality} shows that QAA selection is strongest in the thermoelectric benchmark. Across all thermoelectric policies and evaluation depths, the QAA-selected checkpoint incurs small regret relative to the retrospective-best checkpoint, at or below \(1.8\) points at CP@250 and no more than \(6.2\) points at CP@50. This matches the cleaner saturation pattern visible in Figure~\ref{fig:qaa_super_panel}: thermoelectric performance often reaches an early useful region and then gradually declines.

The COVID-19 benchmark is harder. QAA does not always select near the retrospective-best CP@50 checkpoint, especially for uniform and \(\alpha=1\), where later local peaks produce larger regret. However, regret remains small at broader CP@250 depth across all policies (at most \(3.24\) points), even though CP@50 regret is larger where late local peaks appear. Thus, QAA is strongest as a forecast-time signal for broad candidate recovery: it consistently selects low-regret regimes at CP@250 (at most \(3.24\) points across all policies), while sharp top-50 selection remains more challenging, particularly on COVID-19 where late local peaks produce larger CP@50 regret. This holds even though QAA uses no post-prediction discovery labels.
These results show that QAA tracks the oracle-best checkpoint well. They do not yet show that QAA is \emph{necessary}: a simpler forecast-time rule might work just as well. The most natural such rule is to reuse the target's known pre-prediction positives not as anchors but as a validation set, selecting the checkpoint that best recovers them. We test this alternative next.

\begin{table*}[t]
\centering
\small
\setlength{\tabcolsep}{2.8pt}
\renewcommand{\arraystretch}{1.12}
\resizebox{\linewidth}{!}{%
\begin{tabular}{lcl cccc rr}
\hline
Benchmark & \(K\) & Policy & Direct & Capped & Uncapped & Best (oracle) & \(\Delta\) Capped & \(\Delta\) Uncapped \\
\hline
Thermo & 50  & Uniform & 79.00 {\scriptsize $\pm$0.75} & 76.20 {\scriptsize $\pm$1.57} & 69.40 {\scriptsize $\pm$1.18} & 83.00 {\scriptsize $\pm$1.01} & -2.80 & -9.60 \\
Thermo & 250 & Uniform & 50.52 {\scriptsize $\pm$0.38} & \textbf{53.56} {\scriptsize $\pm$0.46} & \textbf{53.60} {\scriptsize $\pm$0.56} & 51.08 {\scriptsize $\pm$0.47} & +3.04 & +3.08 \\
Thermo & 50  & \(\alpha=1\) & 77.00 {\scriptsize $\pm$1.01} & 76.80 {\scriptsize $\pm$1.54} & 74.20 {\scriptsize $\pm$1.71} & 79.20 {\scriptsize $\pm$0.74} & -0.20 & -2.80 \\
Thermo & 250 & \(\alpha=1\) & 49.00 {\scriptsize $\pm$0.34} & 49.52 {\scriptsize $\pm$0.42} & 48.64 {\scriptsize $\pm$0.51} & 50.12 {\scriptsize $\pm$0.60} & +0.52 & -0.36 \\
Thermo & 50  & \(\alpha\to\infty\) & 58.00 {\scriptsize $\pm$1.78} & 60.20 {\scriptsize $\pm$2.38} & \textbf{65.60} {\scriptsize $\pm$2.22} & 64.20 {\scriptsize $\pm$0.81} & +2.20 & +7.60 \\
Thermo & 250 & \(\alpha\to\infty\) & 44.40 {\scriptsize $\pm$0.30} & 45.48 {\scriptsize $\pm$0.43} & \textbf{46.20} {\scriptsize $\pm$0.45} & 46.16 {\scriptsize $\pm$0.31} & +1.08 & +1.80 \\
\hline
COVID & 50  & Uniform & 37.40 {\scriptsize $\pm$2.87} & 41.80 {\scriptsize $\pm$2.07} & 45.20 {\scriptsize $\pm$2.88} & 52.00 {\scriptsize $\pm$2.61} & +4.40 & +7.80 \\
COVID & 250 & Uniform & 31.52 {\scriptsize $\pm$1.09} & 33.44 {\scriptsize $\pm$1.02} & \textbf{35.32} {\scriptsize $\pm$0.54} & 34.76 {\scriptsize $\pm$0.96} & +1.92 & +3.80 \\
COVID & 50  & \(\alpha=1\) & 37.00 {\scriptsize $\pm$2.26} & 41.80 {\scriptsize $\pm$2.38} & 45.00 {\scriptsize $\pm$2.72} & 48.00 {\scriptsize $\pm$2.52} & +4.80 & +8.00 \\
COVID & 250 & \(\alpha=1\) & 32.92 {\scriptsize $\pm$0.86} & 34.44 {\scriptsize $\pm$0.87} & \textbf{35.92} {\scriptsize $\pm$0.96} & 35.60 {\scriptsize $\pm$0.70} & +1.52 & +3.00 \\
COVID & 50  & \(\alpha\to\infty\) & 33.80 {\scriptsize $\pm$2.07} & 38.20 {\scriptsize $\pm$1.71} & 41.40 {\scriptsize $\pm$2.44} & 44.20 {\scriptsize $\pm$0.81} & +4.40 & +7.60 \\
COVID & 250 & \(\alpha\to\infty\) & 28.44 {\scriptsize $\pm$0.78} & \textbf{30.36} {\scriptsize $\pm$0.51} & \textbf{33.48} {\scriptsize $\pm$0.68} & 30.24 {\scriptsize $\pm$0.62} & +1.92 & +5.04 \\
\hline
\end{tabular}%
}
\caption{Anchor-supported reranking at QAA-selected checkpoints. \(K\) denotes CP@\(K\). Direct is the query-based ranking at the QAA-selected checkpoint; Capped and Uncapped are the two reranking variants; these three are the options available at forecast time. Best (oracle) is the highest cumulative precision the direct ranking attains anywhere in the walk sweep, chosen using post-prediction labels, so it is a hindsight upper bound, not a deployable method. Because reranking changes the ranking itself rather than only the checkpoint, a deployable method can exceed it. Bold marks deployable scores that reach or exceed the oracle. This occurs across CP@250, showing the method is most effective for broad candidate recovery; at CP@50 it holds only under the entity-focused \(\alpha\to\infty\) policy, reflecting that sharp top-of-ranking selection is harder. \(\Delta\) is the percentage-point change relative to Direct. Values are seed means with 95\% CIs (Section~\ref{sec:experimental_setup}).}
\label{tab:qaa_reranking}
\end{table*}

\begin{figure*}[!t]
    \centering
    \includegraphics[
        width=\textwidth,
        height=0.78\textheight,
        keepaspectratio
    ]{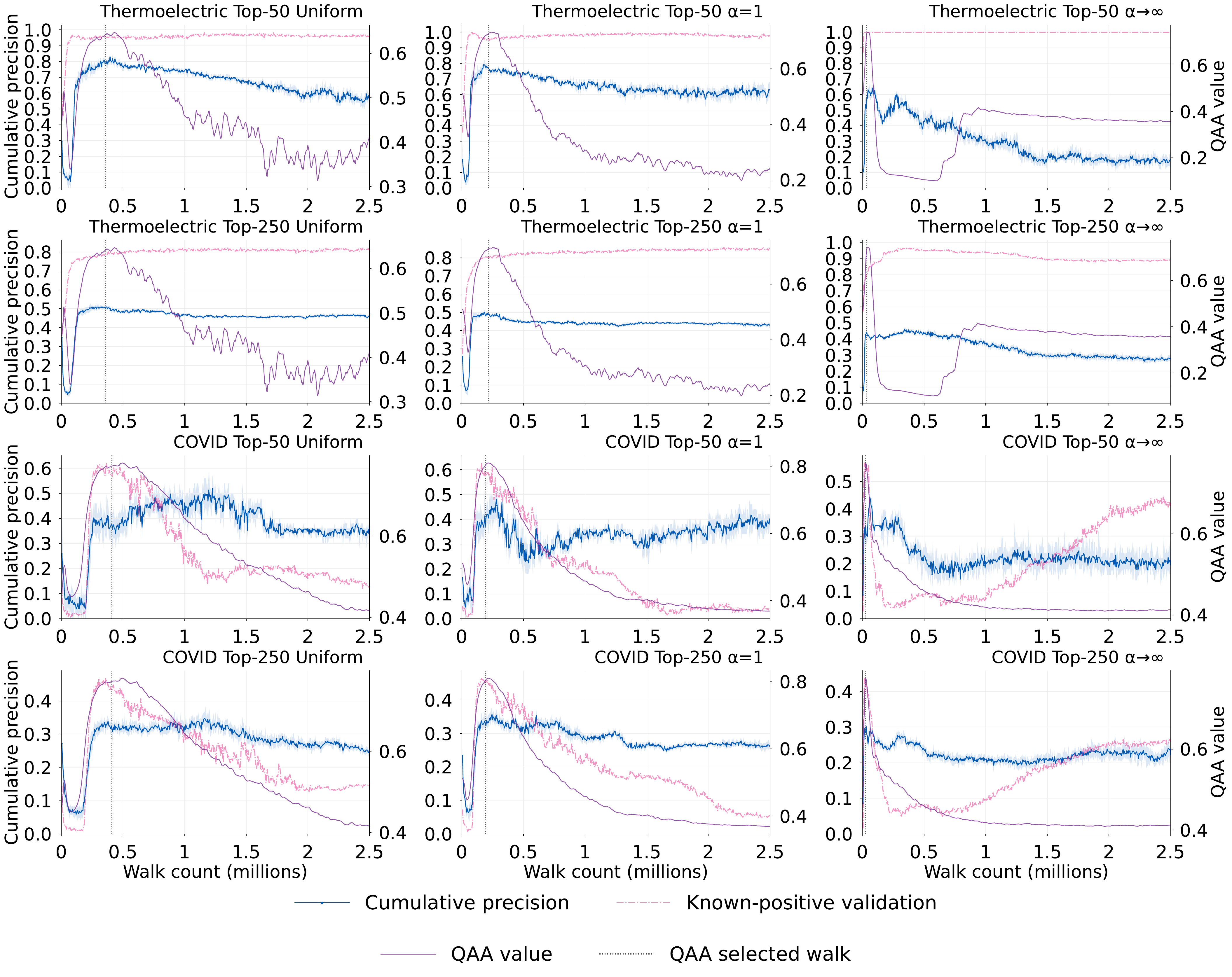}
    \caption{Cumulative precision (blue), QAA (purple), and known-positive validation precision (pink, right axis) across usable-corpus checkpoints, for each benchmark, evaluation depth, and walk policy. Known-positive validation ranks candidates by how well they recover the target's pre-prediction positives, a future-label-free rule available at forecast time. Vertical dotted lines mark the QAA-selected checkpoint. QAA value and known-positive validation are plotted and scaled on the same y-axis. QAA plateaus within the high-precision region, whereas known-positive validation does not track forecast quality: it saturates near its ceiling for thermoelectrics (never signaling when to stop) and peaks early then collapses for COVID-19, selecting checkpoints that miss the precision peak.}
    \label{fig:qaa_validation_supergraph}
\end{figure*}

\subsection{Known-Positive Recovery Does Not Identify the Forecasting Regime}

A tempting future-label-free alternative to QAA is to treat the target's known pre-prediction positives as a validation set: at each checkpoint, rank candidates and measure how many known positives are recovered, then select the checkpoint with the best recovery. Like QAA, this rule uses no post-prediction discovery labels, so it is available at forecast time. Unlike QAA, it selects for reproducing the historical target neighborhood rather than for internal query--anchor stability. Figure~\ref{fig:qaa_validation_supergraph} overlays this known-positive validation signal (pink) on cumulative precision (blue) and QAA (purple) across the full sweep.

The known-positive validation signal does not identify the useful forecasting regime, and it fails in two distinct ways. In the thermoelectric benchmark, known-positive recovery rises quickly and then saturates near its ceiling: under \(\alpha\to\infty\) it reaches essentially perfect recovery and stays there, and under uniform and \(\alpha=1\) it climbs to roughly \(0.98\)--\(1.00\) and remains there through the end of the sweep, with its maximum falling near \(1.1\)--\(1.7\) million walks. Cumulative precision, by contrast, peaks much earlier---near \(0.08\)--\(0.4\) million walks for these policies---and then declines. A practitioner following the known-positive rule would therefore keep adding walks far past the useful region, precisely into the regime where forecast quality degrades. The signal never says stop.

In the COVID-19 benchmark the failure mode is reversed. Known-positive recovery does exhibit a clear interior peak, but it peaks early---near \(0.15\)--\(0.37\) million walks---and then collapses back toward its starting value, whereas cumulative precision peaks later and elsewhere. Selecting on known-positive recovery would place the checkpoint in a region that does not correspond to the cumulative-precision peak, and the sharp post-peak collapse would make the choice unstable across policies. In neither benchmark does the checkpoint favored by known-positive recovery align with the checkpoint that maximizes forecast quality.

QAA behaves differently because it measures relative stability rather than historical recovery. It plateaus within the useful cumulative-precision regime---at \(35\)k--\(355\)k walks for thermoelectrics and \(25\)k--\(410\)k for COVID-19---instead of climbing toward perfect historical recovery or collapsing after an early peak. Recovering known positives measures familiarity with what is already known, whereas QAA tracks whether the query-based and anchor-induced rankings have stabilized relative to each other, which is why it aligns with forecast quality and known-positive recovery does not. The comparison with the oracle-best checkpoint in Table~\ref{tab:qaa_selection_quality} confirms that this is the more useful forecast-time criterion.

\subsection{Anchor-Supported Reranking Can Recover Performance within QAA-Selected Regimes}

QAA selects a saturated walk regime, but it does not directly change the candidate ordering, and the selected checkpoint is not always the retrospective precision maximum. This gives anchor-supported reranking a natural role. When query--anchor agreement is high and saturated, anchor proximity is more likely to reflect meaningful support rather than noise, so reordering candidates by anchor support at the QAA-selected checkpoint can recover performance lost from imperfect checkpoint placement and improve ordering within the selected regime.

Table~\ref{tab:qaa_reranking} shows that anchor-supported reranking can substantially recover this gap.

The strongest reranking gains appear at broader evaluation depth. At CP@250, anchor-supported reranking is highly effective: uncapped reranking produces the highest precision in five of the six benchmark--policy settings, in several cases matching or exceeding the hindsight oracle. In COVID-19, this pattern holds for all three policies. Uniform improves from 31.52\% direct CP to 35.32\% after uncapped reranking, rivaling the oracle value of 34.76\% within confidence intervals. The same pattern appears for \(\alpha=1\), where uncapped reranking reaches 35.92\% compared with an oracle value of 35.60\%, and for \(\alpha\to\infty\), where uncapped reranking reaches 33.48\%, clearly exceeding the oracle value of 30.24\%. These results show that, once QAA has selected a target-coherent embedding regime, anchor support can substantially improve the ordering of a broader candidate region.
\par

The CP@50 results are more sensitive, but they are not broadly negative. Reranking improves CP@50 in four of the six benchmark--policy settings, with the largest gains appearing when the direct QAA-selected ranking leaves more room for recovery. In COVID-19, reranking improves Top-50 performance for all three policies, with uncapped gains of 7.80 points for uniform, 8.00 points for \(\alpha=1\), and 7.60 points for \(\alpha\to\infty\). In thermoelectrics, reranking is less useful for the already-strong uniform and \(\alpha=1\) Top-50 rankings, where direct QAA-selected precision is already close to the oracle. The clearest Thermo Top-50 gain occurs for \(\alpha\to\infty\), where uncapped reranking improves CP@50 from 58.00\% to 65.60\%, exceeding the oracle.

The results bear out the risk--reward tradeoff described in Section~\ref{sec:reranking}. Uncapped reranking produces the largest gains and accounts for most cases where reranking exceeds the oracle, but it also causes the largest loss when it overrides an already strong query ranking (Thermo CP@50, uniform). Capped reranking gains less on average but bounds this downside.

The main conclusion is that QAA and anchor-supported reranking are complementary. QAA identifies a stable regime, and reranking then reuses the same anchor structure to reorder candidates within it. The effect is strongest at CP@250, where the goal is to improve a broader candidate region rather than only the very top of the ranking. This supports the interpretation of QAA as more than a stopping rule: once a stable query--anchor regime has been identified, the same anchor structure can provide useful signal for forecast refinement.

These gains suggest that anchor support is not merely duplicating the query ranking. Instead, anchors appear to identify candidate regions that the query ranking alone does not fully prioritize. To test this directly, we next compare candidates supported by both the query and anchors with candidates supported by the query ranking alone.

\begin{figure*}[!t]
    \centering
    \includegraphics[
        width=\textwidth,
        height=0.78\textheight,
        keepaspectratio
    ]{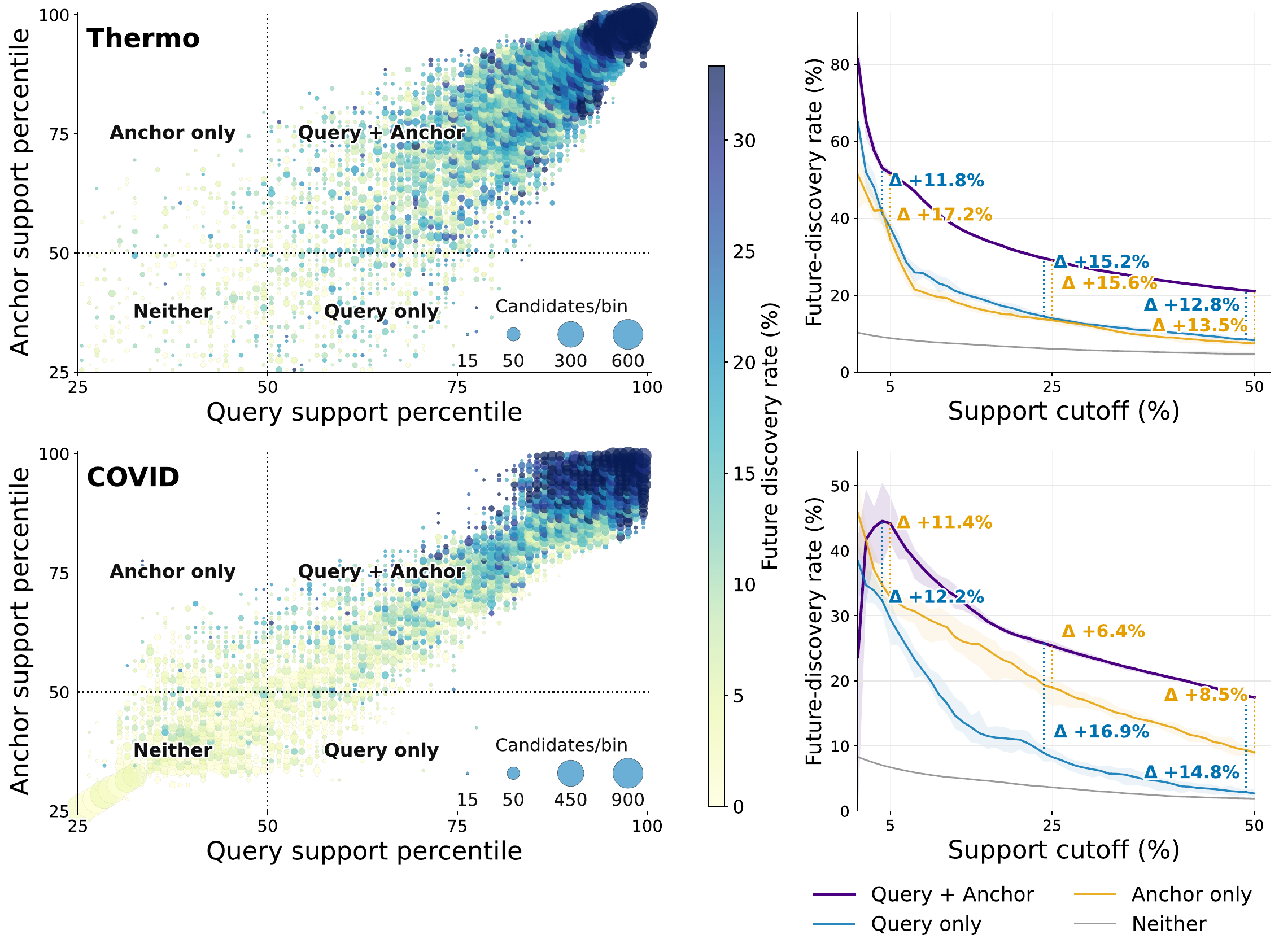}
    \caption{Left panels show the query--anchor support landscape at QAA-selected checkpoints: color indicates future-discovery rate and marker size indicates candidates per bin; bins with fewer than 15 candidates are omitted. Higher support percentiles indicate stronger query or anchor support, so the upper-right region corresponds to candidates supported by both signals. Right panels show future-discovery rate across support cutoffs for Query+Anchor, Query-Only, Anchor-Only, and Neither groups, averaged across the three walk policies for each benchmark; shaded areas indicate the seed min--max range across ten seeds. Comparison annotations are placed at slightly offset cutoffs for legibility, and differences are in percentage points. At the reported cutoffs, Query+Anchor candidates are the most discovery-dense group in both Thermo and COVID.}
    \label{fig:qaa_candidate_support_spectrum}
\end{figure*}

\subsection{Candidate-Level Query--Anchor Support Enriches Future Discoveries}

Having established that the QAA-selected checkpoint marks a stable regime and that anchor-supported reranking improves cumulative precision, we now ask \emph{why} reranking helps. Concretely, we test whether candidates that are supported by both the query and the anchors are more discovery-dense than candidates supported by the query alone. If anchors merely duplicated the query signal, adding anchor support would not identify any additional discovery-rich candidates; if instead anchors carry complementary information, candidates that are jointly query- and anchor-supported should be enriched for future discoveries relative to query-only candidates. Figure~\ref{fig:qaa_candidate_support_spectrum} summarizes this analysis: the left panels place candidates in the query--anchor support landscape at the QAA-selected checkpoint, and the right panels track the future-discovery rate of the four support groups (Query+Anchor, Query-Only, Anchor-Only, Neither) across support cutoffs, averaged over the three walk policies and ten seeds per benchmark. Unless noted, the rates quoted below are taken at the $5\%$ support cutoff, $\rho = 0.05$ (top $5\%$ of each support signal).

The left panels of Figure~\ref{fig:qaa_candidate_support_spectrum} are the most direct view of this test. Each candidate is placed by its query-support percentile (horizontal axis) and its anchor-support percentile (vertical axis), so position encodes \emph{how} a candidate is supported: the lower-left corner holds candidates weakly favored by both signals, the lower-right and upper-left edges hold candidates favored by only one signal, and the upper-right corner holds candidates favored jointly by the query and the anchors. Bubble color encodes the future-discovery rate and bubble size encodes how many candidates fall in each bin, with sparse bins (fewer than 15 candidates) omitted so that the color field reflects stable estimates rather than noise. Read this way, the panel makes the mechanism visible as a spatial gradient rather than a single number: the discovery rate is not uniform across the landscape but deepens systematically as one moves toward the upper-right, where query and anchor support coincide. This is what makes the landscape view powerful---it shows that the two signals are not interchangeable coordinates but complementary axes, and that their agreement, not either one alone, marks the region where future discoveries concentrate. Crucially, this is exactly the corner that anchor-supported reranking promotes, so the left panels give a geometric picture of why reranking moves precision in the right direction.

The right panels turn this spatial picture into a quantitative comparison. For each support cutoff $\rho$, candidates are split into the four groups and their future-discovery rates are plotted, so the curves ask directly whether the jointly supported corner is denser than the single-signal edges across the reported cutoffs. At the $5\%$ support cutoff, $\rho = 0.05$, this comparison is sharpest, and the values quoted below confirm what the left-panel gradient shows: the Query+Anchor group is the densest at the reported cutoffs.

Query+Anchor candidates are the most discovery-dense group in both benchmarks. In Thermo, the future-discovery rate rises from a pool baseline of $0.111$ to $0.517$ for Query+Anchor candidates, versus $0.375$ for Query-Only, $0.345$ for Anchor-Only, and $0.088$ for Neither. This corresponds to a $4.66\times$ enrichment of Query+Anchor over the pool baseline, and a $1.38\times$ gain over the Query-Only group. In COVID, the future-discovery rate rises from a pool baseline of $0.0895$ to $0.442$ for Query+Anchor candidates, versus $0.295$ for Query-Only, $0.328$ for Anchor-Only, and $0.067$ for Neither, a $4.94\times$ enrichment over the pool and a $1.50\times$ gain over Query-Only. In both cases, at this cutoff, the joint Query+Anchor group is the densest, and its rate exceeds that of query support alone.

This pattern supports the interpretation that reranking benefits from promoting jointly supported candidates. Query+Anchor candidates are consistently more discovery-dense than Query-Only candidates, which is consistent with anchors adding signal rather than restating what the query already encodes, and with anchor support identifying discovery-rich regions that the query ranking alone does not fully prioritize. This is the quantitative counterpart of the upper-right gradient in the left panels of Figure~\ref{fig:qaa_candidate_support_spectrum}, and is consistent with reranking helping by reweighting candidates toward this jointly supported region.

Two caveats temper this reading. First, the Query+Anchor group is small---on the order of $180$ candidates in Thermo and $60$ in COVID at $\rho = 0.05$---so its rates, while consistently the highest, are estimated from a limited number of candidates; the shaded bands in Figure~\ref{fig:qaa_candidate_support_spectrum} report the corresponding seed-to-seed spread. Second, the ranking of the two single-signal groups is not identical across benchmarks: Query-Only edges out Anchor-Only in Thermo, whereas Anchor-Only is slightly denser than Query-Only in COVID, so we do not claim a universal ordering between the single-signal groups. What is robust is that combining both signals is best in both benchmarks.

%% file: sections/discussions.tex
\paragraph{Forecast-time readiness}
The most useful outcome of our study is that the stability of a discovery forecast can be assessed at the moment the forecast is issued, without waiting for future discoveries to reveal whether the forecast was correct. This matters for prospective use. A forecaster must choose sampling budgets and decide which candidate rankings to act on before any future labels exist, whereas true accuracy, novelty, and impact can only be measured retrospectively. QAA tests representational readiness directly in the embedding used for forecasting, requiring only information available before the prediction date: the trained representation, the target query, and known pre-prediction positives used as anchors.

What we find most interesting is that this can be done without holding out any data at deployment. A live forecasting system should generally use all data available up to the prediction date, since holding out recent history creates a cleaner validation problem but leaves the deployed forecast less current. QAA avoids that tradeoff by asking whether the target query and its anchors induce consistent candidate neighborhoods within the full pre-prediction representation. In our experiments, future discoveries are held out only to evaluate the diagnostic after forecasts are produced; they are never used by the diagnostic itself. Across the thermoelectric and COVID-19 benchmarks, QAA-selected walk-count regimes rival those chosen retrospectively with future labels, and anchor-supported reranking within those regimes matches or exceeds hindsight-tuned forecasts for broad candidate recovery (CP@250); at the very top of the ranking (CP@50), reranking still recovers substantial precision on COVID-19 even though selection alone is harder there. Representation readiness can therefore be assessed prospectively, rather than only judged after future labels have appeared.

It is worth being clear about the scope of this result. QAA is not an estimator of a candidate's discovery probability, eventual novelty, or impact. It tests whether the query-relevant geometry of a learned random-walk representation has stabilized with respect to known pre-prediction positives. We view this narrowness as a strength, because readiness is one of the few properties of a discovery forecast that can actually be assessed at forecast time.

\begin{table*}[t]
\centering
\small
\setlength{\tabcolsep}{2.4pt}
\renewcommand{\arraystretch}{1.12}
\resizebox{\linewidth}{!}{%
\begin{tabular}{lc ccc cc ccc}
\hline
& & \multicolumn{5}{c}{At 250k} & \multicolumn{3}{c}{QAA RRF (plateau)} \\
\cline{3-7} \cline{8-10}
Benchmark & \(K\) & Uniform & \(\alpha=1\) & \(\alpha\to\infty\) & S2 & S3 & Query-only & Capped & Uncapped \\
\hline
Thermo & 50  & 75.00 {\scriptsize $\pm$1.82} & 75.80 {\scriptsize $\pm$0.45} & 53.20 {\scriptsize $\pm$1.00} & 46.00 & 46.00 & 73.80 {\scriptsize $\pm$0.45} & \textbf{77.80} {\scriptsize $\pm$1.06} & 76.60 {\scriptsize $\pm$2.14} \\
Thermo & 250 & \textbf{51.00} {\scriptsize $\pm$0.59} & 48.68 {\scriptsize $\pm$0.45} & 43.40 {\scriptsize $\pm$0.41} & 33.20 & 34.80 & 48.52 {\scriptsize $\pm$0.33} & 50.36 {\scriptsize $\pm$0.25} & 49.20 {\scriptsize $\pm$0.57} \\
\hline
COVID & 50  & 33.20 {\scriptsize $\pm$1.81} & 44.40 {\scriptsize $\pm$3.15} & 33.40 {\scriptsize $\pm$2.03} & 40.00 & 40.00 & 32.80 {\scriptsize $\pm$1.68} & 38.60 {\scriptsize $\pm$1.36} & \textbf{47.40} {\scriptsize $\pm$1.36} \\
COVID & 250 & 28.00 {\scriptsize $\pm$1.21} & 34.04 {\scriptsize $\pm$0.68} & 25.20 {\scriptsize $\pm$0.57} & 32.80 & 33.20 & 31.80 {\scriptsize $\pm$0.72} & 33.40 {\scriptsize $\pm$0.61} & \textbf{35.20} {\scriptsize $\pm$1.15} \\
\hline
\end{tabular}%
}
\caption{Cumulative precision CP@\(K\) (percent, seed mean \(\pm\) 95\% CI where applicable). Single walk policies are at a fixed 250{,}000-walk budget; the S2/S3 baselines of \citet{sourati2023accelerating} are fixed graph-transition baselines that do not vary with the walk budget; cross-policy RRF fusion is at its QAA-selected plateau. Query-only is the cross-policy fusion of the three policies' query-based rankings at their respective QAA-selected checkpoints, with no anchor contribution; Capped and Uncapped add anchor support to that fusion, capped at or unrestricted by the query contribution respectively. Bold values mark the best method in each row. QAA-selected cross-policy fusion is competitive with or better than any single walk policy fixed at 250{,}000 walks, tracking the strongest single policy on thermoelectrics and giving the best score on both COVID-19 depths, without committing in advance to a walk budget or policy.}
\label{tab:crosspolicy_fusion}
\end{table*}

\paragraph{Benchmark limits}
The two benchmarks also bound what QAA, and the underlying forecasting framework, can claim, and the clearest limitations we observed come from the datasets themselves. The two tasks define ground-truth discovery differently. Thermoelectric positives are first-time co-occurrences between a material and a property keyword, a permissive signal that need not imply a confirmed physical property, whereas COVID-19 positives are drugs advanced to clinical trials, a stricter and more costly form of evidence. Precision is therefore not directly comparable across tasks, and thermoelectric positives should be read as literature associations rather than validated discoveries.

The framework can also only rank candidates already present in the pre-prediction graph. This is a substantial restriction in the thermoelectric task, where only \(828\) of the \(3{,}720\) future positives fall inside the scored candidate pool, so most eventual positives are structurally outside the benchmark's scoring universe. QAA assesses readiness for the forecast the representation can actually make; it cannot recover discoveries that the underlying graph gives the model no path to reach.

The COVID-19 hypergraph is also far more author-dominated than the thermoelectric one: in the drug-repurposing setting author nodes outnumber drug nodes by more than a thousand to one (over \(99.95\%\) versus under \(0.05\%\)), compared with roughly \(94\%\) versus \(6\%\) for the energy-materials setting~\citep{sourati2023accelerating}. Under this extreme imbalance, walks recruit far fewer candidate-entity co-occurrences per unit of sampling, which helps explain why the COVID-19 QAA and cumulative-precision curves are noisier and why plateau selection is less sharp there, especially at CP@50.

\paragraph{Policy choice and robustness}
A further practical difficulty is that no walk policy is uniformly best. In our sweeps the strongest policy depends on both the benchmark and the recovery depth. Uniform sampling is strongest on thermoelectric recovery, where the entity-focused \(\alpha\to\infty\) policy is markedly weaker, while on COVID-19 the ordering shifts and \(\alpha\to\infty\) becomes competitive. Because the best policy is not known before outcomes are observed, choosing a policy in advance can itself become an implicit source of retrospective selection.

As an exploratory robustness analysis, we combined the three policies at their QAA-selected plateaus using reciprocal-rank fusion. We do not present fusion as a primary contribution, and it does not uniformly outperform the best single policy. Instead, it illustrates a practical way to obtain strong performance without having to guess which policy will work, since different policies perform well on different tasks and at different recovery depths. As Table~\ref{tab:crosspolicy_fusion} shows, QAA-selected fusion is competitive with or better than the single walk policies fixed at the arbitrary 250{,}000-walk budget: fused rankings track the best single policy on the thermoelectric task, where uniform sampling is strongest and \(\alpha\to\infty\) is weak, and match or exceed every single policy at both COVID-19 recovery depths, though the margin is narrow at CP@250. Crucially, fusion attains this without committing to an arbitrary walk budget or a walk-policy choice, since QAA selects the checkpoint and reciprocal-rank fusion removes the need to pick a single policy. Fusion therefore trades a small amount of peak thermoelectric performance for greater robustness across tasks and policies.

\paragraph{Where the forecasting signal lives}
Finally, our results raise a question about where the useful forecasting signal resides. The \(\alpha\to\infty\) policy suppresses author-mediated traversal, yet remains useful and is competitive on COVID-19. This suggests that a substantial part of the near-term forecasting signal may lie in the relational structure among entities, properties, and biomedical concepts themselves, rather than only in the distribution of expert attention.

A natural direction for future work is to move beyond publication co-occurrence and model scientific relations more directly, using curated interaction networks, structured property databases, or explicitly extracted relational statements. Such representations may expose paths unavailable in publication-derived hypergraphs and may better distinguish literature association from substantive discovery. The open question is whether QAA-style readiness diagnostics transfer to these richer relational settings, where the same forecast-time problem remains: determining whether a learned representation is stable enough to support a prospective forecast before future outcomes are known.

%% file: sections/conclusion.tex
We introduced Query--Anchor Agreement (QAA), a future-label-free diagnostic that requires no separately trained model and assesses whether a random-walk discovery-forecasting representation has stabilized enough to support a forecast at the moment the forecast must be made. Rather than estimating the probability, novelty, or impact of a future discovery, QAA reuses the learned embedding to check whether a target query and its known pre-prediction positives retrieve overlapping regions of the candidate pool, and identifies the walk-count regime within a policy where this agreement saturates. Across the thermoelectric and COVID-19 benchmarks, QAA selects regimes competitive with retrospectively chosen settings, and QAA-guided anchor reranking matches or exceeds hindsight-tuned forecasts for broad candidate recovery---while sharp top-rank precision, especially COVID-19 CP@50, remains harder---all without using future discovery labels. By separating the question of representational readiness from the question of eventual correctness, QAA lets random-walk discovery forecasts be evaluated prospectively rather than only in hindsight, and points toward readiness diagnostics for the richer, more direct relational representations that future discovery-forecasting systems are likely to use.

%% file: sections/credit.tex
\textbf{Siu Tin Wu:} Conceptualization, Methodology, Software, Formal analysis, Investigation, Visualization, Writing -- original draft, Writing -- review \& editing.
\textbf{Ryan C. Barron:} Software, Writing -- review \& editing.
\textbf{Phoebe Upton Sanderbeck:} Validation, Writing -- review \& editing.
\textbf{Kim {\O.} Rasmussen:} Funding acquisition, Supervision, Writing -- review \& editing.
\textbf{George Tompkins:} Supervision, Writing -- review \& editing.
\textbf{Maksim E. Eren:} Software, Supervision, Project administration, Writing -- review \& editing.

%% file: sections/competing_interests.tex
The authors declare that they have no known competing financial interests or personal relationships that could have appeared to influence the work reported in this paper.

%% file: sections/data_availability.tex
No new data were generated in this study. The thermoelectric-materials and COVID-19 drug-repurposing benchmarks analyzed in this work were released by \citet{sourati2023accelerating} and are publicly available from the original authors.

%% file: sections/acknowledgment.tex
This manuscript has been approved for unlimited release and has been assigned LA-UR-26-26875. Funding: This work was supported by the U.S. Department of Energy National Nuclear Security Administration's Office of Defense Nuclear Nonproliferation Research and Development (DNN R\&D) [Contract No. 89233218CNA000001]. LANL is operated by Triad National Security, LLC, for the National Nuclear Security Administration of the U.S. Department of Energy. The funders had no role in study design, data collection and analysis, decision to publish, or preparation of the manuscript.

%% file: sections/ai_declaration.tex
During the preparation of this work the author(s) used Claude for editorial assistance (language, structure, and clarity). The author(s) reviewed and edited all content and take full responsibility for the content of the published article.